\documentclass[twoside,11pt]{article}
\pdfoutput=1
\usepackage{jmlr2e}
\usepackage{subcaption}
\usepackage{mathtools}  
\usepackage{comment}
\usepackage[boxed,figure]{algorithm2e}
\usepackage{xcolor}
\usepackage{setspace}
\usepackage[noabbrev]{cleveref}
\DeclareMathOperator*{\argmax}{arg\,max}
\DeclareMathOperator*{\argmin}{arg\,min}

\newcommand{\km}{{\ttfamily k-means}} 
\newcommand{\kmp}{{\ttfamily k-means++}} 
\newcommand{\kmu}{{\ttfamily k-means-u}} 
\newcommand{\kms}{{\ttfamily k-means-u*}} 

\ShortHeadings{The \kms{} algorithm}{B. Fritzke}
\firstpageno{1}

\begin{document}
	
\renewcommand\floatpagefraction{.6}
\title{The \kms{} algorithm: non-local jumps and greedy retries improve \kmp{} clustering}

\author{\name Bernd Fritzke \email fritzke@web.de \\
       Friedrichsdorf\\
       61381 Germany
       }

\editor{}

\maketitle
\setcitestyle{authoryear, open={(},close={)}}
\begin{abstract}

We present a new clustering algorithm called \kms{} which in many cases is able to significantly improve the clusterings found by \kmp{}, the current de-facto standard for clustering in Euclidean spaces. First we introduce the \kmu{} algorithm which starts from a result of \kmp{} and attempts to improve it with a sequence of non-local ``jumps'' alternated by runs of standard \km{}. Each jump transfers the ``least useful'' center towards the center with the largest local error, offset by a small random vector. This is continued as long as the error decreases and often leads to an improved solution. Occasionally \kmu{} terminates despite obvious remaining optimization possibilities. By allowing a limited number of retries for the last jump it is frequently possible to reach better local minima. The resulting algorithm is called \kms{} and dominates \kmp{} wrt.~solution quality which is demonstrated empirically using various data sets. By construction the logarithmic quality bound established for \kmp{} holds for \kms{} as well.
\end{abstract}


\section{Introduction}
In machine learning various sub-fields can be distinguished based on the amount of feedback given to the learner. On one side there is \emph{supervised learning} where the given data consists of pairs of input and corresponding output. In this case the learner is trained to approximately reproduce this mapping with the expectation that it can afterwards generalize to unseen patterns. Typical examples of supervised learning are classification (e.g. image or speech recognition) and regression (e.g. time-series prediction). Considerably less information is given in \emph{reinforcement learning} where a system is expected to learn how to perform long sequences of actions (e.g. in game play) while only receiving occasional feedback (e.g. "game won" or "game lost"). Finally there is \emph{unsupervised learning} where a system is expected to learn purely from unlabeled data. It has been argued recently that due to the number of connections in the human brain and the life span of humans a large part of learning must be unsupervised (G. Hinton, AMA on Reddit.com, 2015). This paper is concerned with \emph{clustering} which is considered to be a fundamental unsupervised learning problem. In particular we propose a method to improve clusterings generated by the \kmp{} algorithm which currently can be seen as a de facto standard for clustering numerical data.

The remainder of this article is organized as follows: In \cref{sec:clustering} the particular kind of clustering problem is described which the new algorithm (as well as \km{} and \kmp) deal with. 

In \cref{sec:km} the classical \km{} algorithm is introduced and illustrated by examples which also highlight its problematic dependency on initialization. 

In section \cref{sec:kmp} we describe the more recent \kmp{} algorithm which enhances \km{} by a stepwise initialization which in general leads to much better results than e.g.~the common random initialization from the data set. 

\Cref{sec:kmpwhy} investigates why  \kmp{} is so effective and shows for a simple example problem that under certain conditions \kmp{} always finds the optimal initial placement. 

\Cref{sec:kmphard} gives examples of problems which are hard for \kmp{} in the sense that \kmp{} is in general not able to find a solution close to the optimum (which is known for these specific problems by construction). It is investigated why this is the case and that for certain examples \kmp{} has difficulties if $k$ is larger than the number of clusters in the data. 

In \cref{sec:kmu} the \kmu{} algorithm is introduced which considerably improves many solutions found by \kmp{} via a sequence of non-local jumps alternated with \km{} phases. 

In \cref{sec:occasional} an occasional problem of \kmu{} is described and illustrated by an example: too early termination due to a poor (but normally still better than \kmp) local minimum. 

\Cref{sec:kms} introduces the \kms{} algorithm which enhances \kmu{} by allowing greedy retries. This allows in many cases to reach considerably better local minima. 

In \cref{sec:empirical} the results of systematic simulations with various data sets are presented covering large ranges of $k$ for each data set. In a nutshell the conclusion is: 
w.r.t.~solution quality \kms{} dominates \kmu{} which again dominates \kmp.

In \cref{sec:guarantees} the fact is pointed out that the logarithmic quality bound which has been established for \kmp{} holds (by construction) for \kms{} as well.

\Cref{sec:summary} provides a summary of the article.

\section{Clustering}\label{sec:clustering}

Clustering in general can be described as the problem of finding groups (clusters) in a data set such that the similarity of data items within a group is large and data items from different groups are dissimilar. 
The used measure of similarity can vary from a subjective perceptual criterion to precise mathematical definitions. In the context of this paper we choose the goal of clustering to be a computational one, in particular the minimization of a distance-based error function:

We assume an integer $k$ and a set of $n$ data points $\mathcal{X} \subset \mathbb{R}^d$. The goal is to select $k$  centers $\mathcal{C}$ such that the error function

\begin{equation}
\phi(\mathcal{C},\mathcal{X}) = \sum_{x\in\mathcal{X}} \min\limits_{c\in\mathcal{C}} ||x-c||^2 \label{eq:1}
\end{equation}
is minimized. We thus strive to position the centers in such a way that the sum of squared distances between each data point and its respective closest center is minimized. In this paper we will refer to $\phi(\mathcal{C},\mathcal{X})$ also as \emph{Summed Squared Error} or SSE.

\begin{figure}
	\centering
	\includegraphics[width=0.99\linewidth]{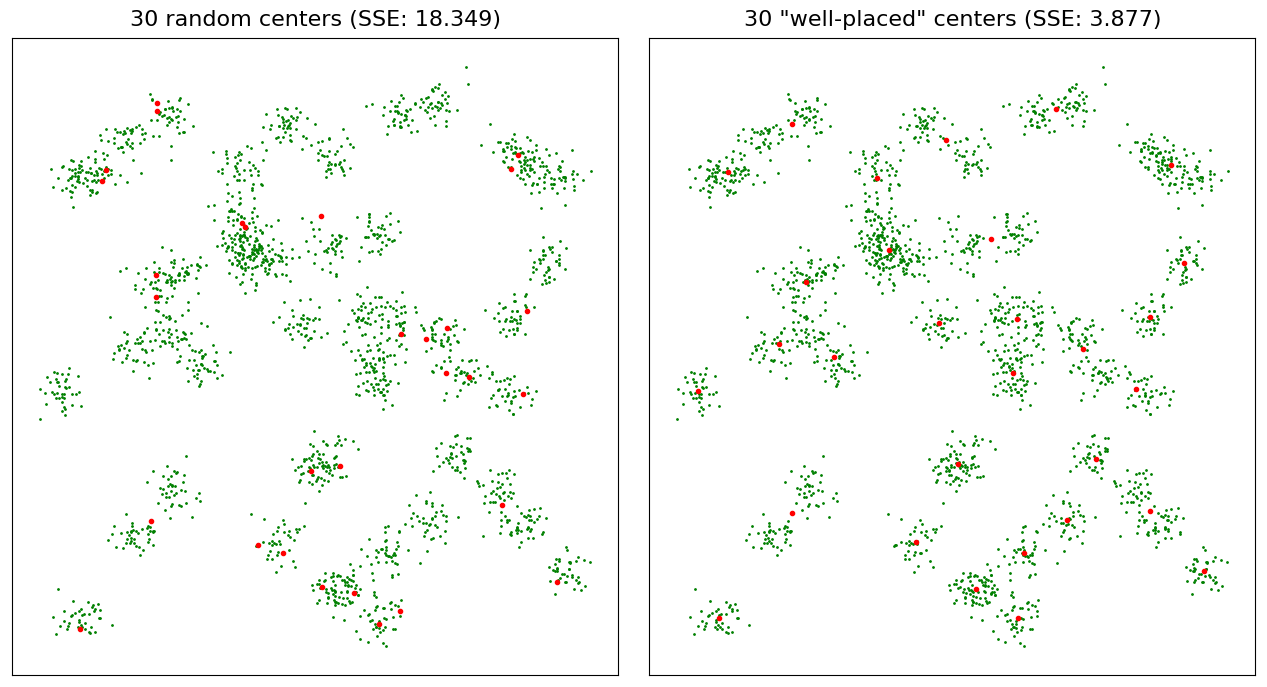}
	\caption[short caption]{Illustration of a poor (left) and a good (right) distribution of centers for a two-dimensional data distribution generated from a mixture of equal-variant Gaussians. The Summed Squared Error (SSE=$\phi$) is considerably smaller for the right distribution of centers.}
	\label{fig:gaussians}
\end{figure}

The resulting set of centers $\mathcal{C}$ can be used to group (cluster) the original data set but also to encode the data for transmission or storage in the sense of vector quantization. Finding the optimal solution for this problem is known to be NP-hard \citep{Aloise2009}, so in practice we need to use approximation algorithms. In \cref{fig:gaussians} an example of a data set is given with two clusterings which vary strongly with respect to the error criterion in \cref{eq:1}.

\section{\km}\label{sec:km}
\km \, \citep{Lloyd1982} is a classical method for clustering or vector quantization. 
Starting from an initial set of centers $\mathcal{C} = \{ c_1,c_2,...\,,c_k\}$ a sequence of so-called \emph{Lloyd Iterations} is performed. Each Lloyd iteration consists of two steps:
%

\begin{itemize}
	\item determine for each center $c_i$ its so-called \emph{Voronoi Set} $C_i$, which is the  set of data points for which $c_i$ is the closest center:
	\begin{equation}\label{eqn:voro}
	C_i := \{x \in \mathcal{X}\,|\; \|x-c_i\| < \|x-c_j\|\, \forall j\neq i\}
	\end{equation} 
	\item move all centers to the center of gravity of their Voronoi set:
			$$c_i := \frac{1}{|C_i|} \sum\limits_{x\in C_i}x$$
\end{itemize}

The complete \km{} algorithm is specified in \cref{alg:km}. It consists of an initialization ("Seeding") followed by a sequence of Lloyd iterations.

\begin{algorithm}[ht]
	\SetKw{ini}{Seeding:}
	\ini Choose $k$ initial centers $\mathcal{C} = \{ c_1,c_2,...\,,c_k\}$\;
	\Repeat
	( /* Lloyd~iterations */ )
	{
		no more change of $\mathcal{C}$
	}{
		\ForEach{
			$i \in \{1, ...,k \}$
		}{
			$C_i\leftarrow \{x \in \mathcal{X}\,|\; \|x-c_i\| < \|x-c_j\|\, \forall j\neq i\}$ 
			\tcc*[r]{$C_i$ is assigned the set of all points in $\mathcal{X}$ having $c_i$ as their closest center}
		}
		\ForEach{
			$i \in \{1, ...,k \}$
		}{
			$c_i \leftarrow \frac{1}{|C_i|} \sum\limits_{x\in C_i}x$
			\tcc*[r]{modify $c_i$ to be the center of mass of $C_i$}
		}
	}
	\Return{$\mathcal{C}$}\;
	\caption{The \km{} algorithm}
	\label{alg:km}	
\end{algorithm}

A common seeding method is to select the initial centers equiprobable at random from the data set $\mathcal{X}$. This approach ensures that there are no unused centers ("dead units") which could occur if centers were selected from random locations not contained in $X$. 

Possible ties in defining $C_i$ can be resolved arbitrarily. If a deterministic resolution method is used, the whole algorithm is deterministic (after the seeding) and leads to reproducible results. In summary the \km{} algorithm performs one Lloyd iteration after another as long as the SSE decreases with each iteration\footnotemark[1].\footnotetext[1]{An alternative stopping criterion is to terminate as soon as the error improvement after any Lloyd iteration falls below a threshold.}

\begin{figure}
	\centering
	\includegraphics[width=0.99\linewidth]{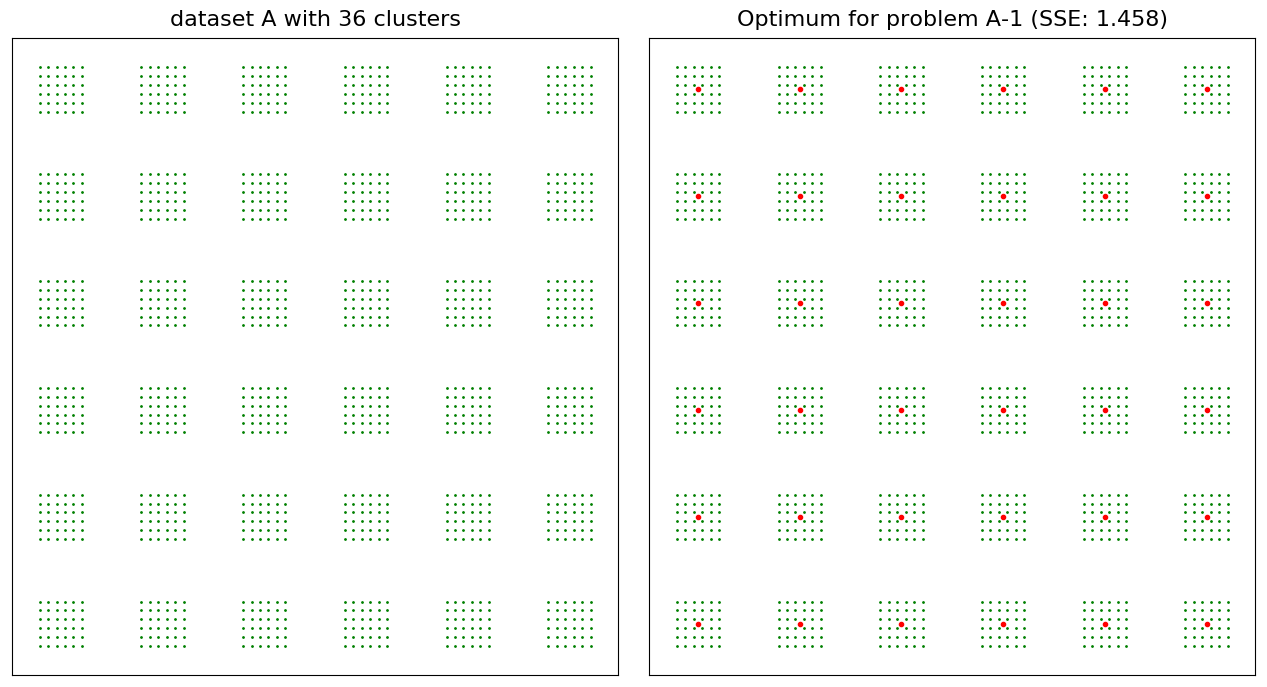}
	\caption[short caption]{left: Data set $A$ consisting of 36 identically-shaped clusters. We denote with problem $A$-1 the task to distribute 36 centers such that the SSE wrt.~$A$ is minimized. Right: Optimal solution for problem $A$-1: one center positioned in each cluster, SSE: 1.458}
	\label{fig:kma1_opt}
\end{figure}

\km{} in this form is very simple and known to converge in a finite number of steps. The quality of its solutions measured by the above error function $\phi$ can however vary strongly depending on the seeding used. Let us illustrate this by applying \km{} to the data set $A$ shown in \cref{fig:kma1_opt} (left). It consists of 36 quadratic clusters. Let us specifically consider the problem to distribute 36 centers over $A$, i.e. the same number as there are clusters in $A$. In the following we denote this particular problem as $A$-1 (variations of this problem where the number of centers is $n$ times as large as the number of clusters in $A$ will accordingly be denoted as $A$-$n$).

Due to the regular structure of the data set and the fact that the number of centers is the same as the number of clusters it is obvious in this case what the optimal arrangement of centers is: One center in each cluster, positioned in the middle of the cluster (see \cref{fig:kma1_opt}, right side). The corresponding summed squared Error (SSE) for this optimal solution is 1.458 (numerically computed).

\begin{figure}
	\centering
	\includegraphics[width=0.99\linewidth]{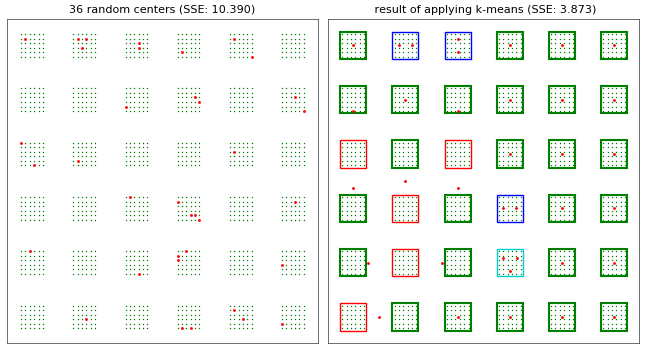}
	\caption[short caption]{\km{} approximately solves clustering problem $A$-1. Left:  $k=36$ centers have been chosen at random from $A$. Right: Result of \km{} applied to the initial configuration shown in the left figure. Many clusters correctly receive one center (optimal, coded green), some clusters however gets 2 or 3 centers (too many, coded dark and light blue) and some clusters do not get any center at all (too few, coded red)}
	\label{fig:kma1}
\end{figure}

Let us now apply \km{} with equiprobable random initialization from the given data set $A$ to this problem. In \cref{fig:kma1} a typical solution is shown. It has a SSE of 3.873 which is 166\% percent above the known optimum (so this solution can be considered rather poor).  

\begin{figure}
	\centering
	\includegraphics[width=0.99\linewidth]{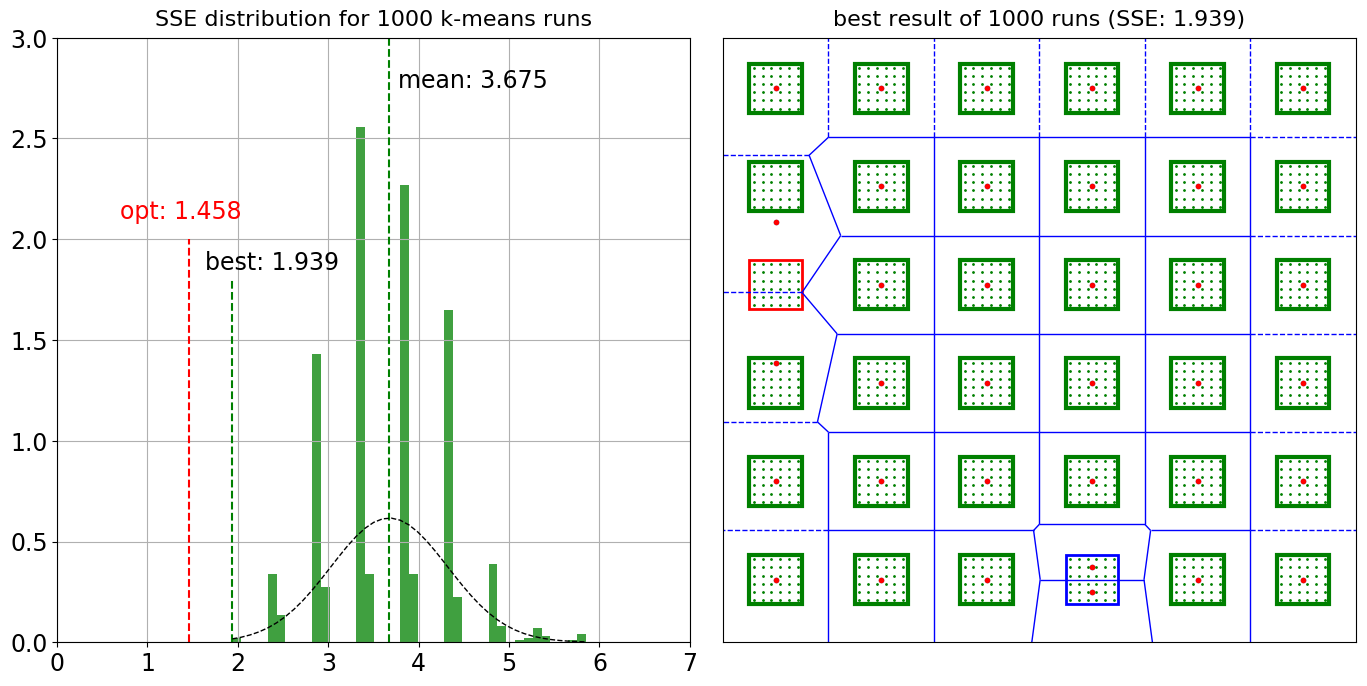}
	\caption[short caption]{ left: normalized distribution of SSE values for 1000 runs, optimal solution and Gaussian fitted to the measured errors. right: best result obtained during the 1000 runs. The SSE of this solution is still 33.0\% higher than that of the known optimum.}
	\label{fig:kma1_brute}
\end{figure}

A rather
obvious way to achieve better results is to run \km{} several times with different random seedings and return the best result of all runs. However, even if we run it 1000 times, the best results are usually still far from the optimum (see \cref{fig:kma1_brute}). If we model the distribution of SSE values obtained from the simulation with a normal distribution, the optimal SSE is more than 3.4$\sigma$ away from the mean indicating an occurrence  probability of less than 0.0003. Even if it is not clear how well a normal distribution can model the \km{} SSE distribution, it seems that a brute-force method based on a large number of random seedings is a costly way to find a low-SSE configuration with \km.

\section{\kmp}\label{sec:kmp}

25 years after the original \km{} algorithm an improvement was proposed by \cite{Arthur2007}, the \kmp{} algorithm which today can be seen as a standard way of doing \km \,(one specific indication being that the  default implementation of \km{} in the popular \texttt{scikit-learn} package \citep{Pedregosa2011a} for scientific computation \emph{is} \kmp). The \kmp{} algorithm augments \km{} by a stepwise seeding phase which takes into account the distance of the data points from the centers placed so far. Specifically the probability that a data point $x$ is selected as the position of the next center is chosen to be proportional to the minimum squared distance to any of the already placed centers. In contrast to choosing the next center equiprobable from the remaining data points this method favors regions which are far from the existing centers. Placing the next center in such a region likely results is a large reduction of the overall error.  The complete algorithm is depicted in \cref{alg:kmp}.

\begin{algorithm}[ht]
	\SetKw{ini}{Initialization:}
	\ini$\mathcal{C}\leftarrow \{c_1\}$ with  $c_1$, chosen uniformly at random from $\cal{X}$\;
	$i\leftarrow 1$\;
	\While(/* choose next center */){$i<k$}{
		$i\leftarrow i+1$\;
		$\mathcal{C}\leftarrow \mathcal{C} \cup \{x\}$ with $x$ drawn at random from $\cal{X}$ with probability $$P(x) = \frac{D(x)^2}{\sum\limits_{x\in\cal{X}}D(x)^2}$$
		whereby $
		D(x) = \min_{c \in \mathcal{C}} \|c-x\|
		$
	}
	Perform \km{} using $\mathcal{C}$ as the initial set of centers\;
	\Return{resulting value of $\mathcal{C}$}\;
	
	\caption{The \kmp{} algorithm}
	\label{alg:kmp}
\end{algorithm}

Notable details in the \texttt{scikit-learn} implementation of \kmp{} are that the algorithm is always run several times (default: 10) and only the best result is returned. Moreover, when placing new centers during seeding several candidates are tried out and the one reducing the overall error most is taken. Since no attempt is made to backtrack this can be seen as a kind of "greedy" algorithm and is already described in \cite{Arthur2007} without further analysis but the remark, that "it helps".

\begin{figure}
	\centering
	\includegraphics[width=1.0\linewidth]{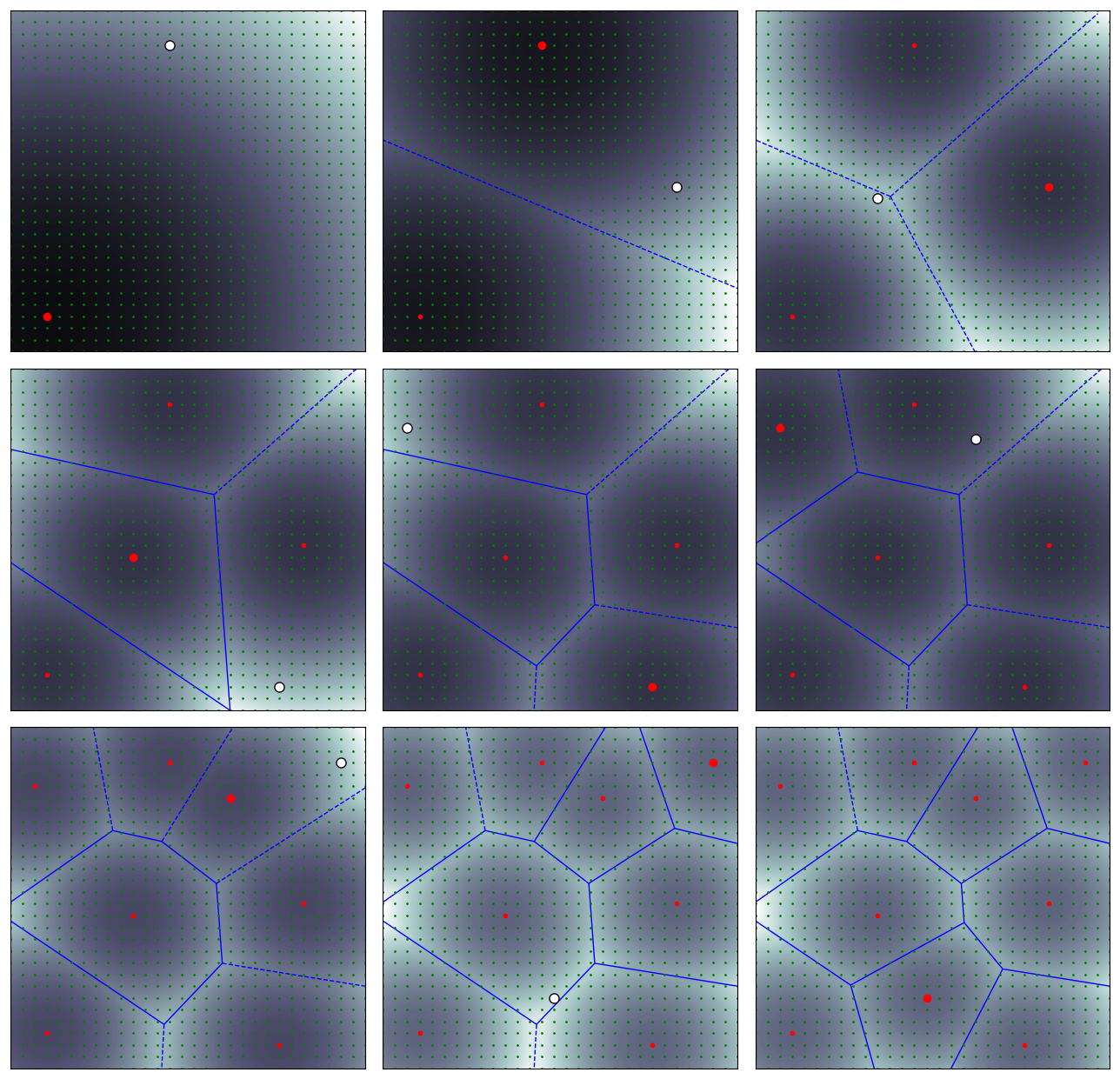}
	\caption[short caption]{\kmp{} seeding example: the data consists of 900 points arranged in a 30x30 grid. The sequence of images shows the insertion of $k=9$ centers depicted as red circles. The most recently inserted center is always enlarged. The shading indicates the squared distance of the data points (dark=low, light=high, but shading of each subfigure is normalized individually to avoid later figures to be all black due to small distances) from the current set of centers and therefore also the probability that \kmp{}  choses the next center from that location. The blue segments show the Voronoi diagram corresponding to the current set of centers. The Voronoi segments coincide with light-colored regions since they are defined by neighboring centers to which they have equal (and thus large) distance at each position. The white dot indicates the position of the next center chosen. It is noticeable that the white dot usually lies in a region which is colored lightly (indicating high distance resp. probability).}
	\label{fig:kmpp_cont}
\end{figure}

In \cref{fig:kmpp_cont} the way \kmp{} works, is illustrated. One can see there how the distance-based probability guides the selection of the respective next center.

\kmp{} is often very effective in finding good seedings for the following \km{} phase. According to \cite{Arthur2007} (but also in accordance with our own experiments) the resulting solutions are mostly significantly better than those obtained by equiprobable random seeding from the data set. There is even a proven lower bound for the quality of the solution $\cal{C}$ constructed by \km{} compared to the optimal solution:

\begin{equation}
E[\phi] \le 8(\mbox{ln}\,k+2)\phi_{\mbox{OPT}}
\end{equation}

For practical purposes this bound may often not be tight enough (for k=100, e.g., the bound guarantees that the error of the solution found differs from the optimum by not more than a factor of 52.8), but the proof of the bound as such is quite remarkable since there is no bound at all for the solutions of \km, i.e. they can be arbitrarily poor.

To illustrate the high quality of \kmp{}  we show - after having performed 400 simulations several times - the two most frequent results for the problem $A$-1 where the original \km{} with random seeding had problems: In 48-50\% of the cases the optimal solution was found (\cref{fig:kma1_opt}). Also in 48-50\% of the cases one cluster remained empty which is actually very good given that for \km{} this was the best solution resulting from 1000 runs and did only occur in about 0.3\% of the simulations (\cref{fig:kma1_brute}). Rarely (0-4\%) two clusters remained empty. One should note here that In every simulation run we performed \kmp{} exactly once. As mentioned above the implementation of \kmp{} in \texttt{scikit-learn} is configured such that per default 10 runs are performed and the best result is returned. Given the described probability of about 50\% for finding the optimal solution of $A$-1, it is trivial to see that the \texttt{scikit-learn} implementation would find the optimal solution with about  $P(opt)=1-(\frac{1}{2})^{10}=0.999$.

\begin{figure}
	\centering
	\includegraphics[width=1.0\linewidth]{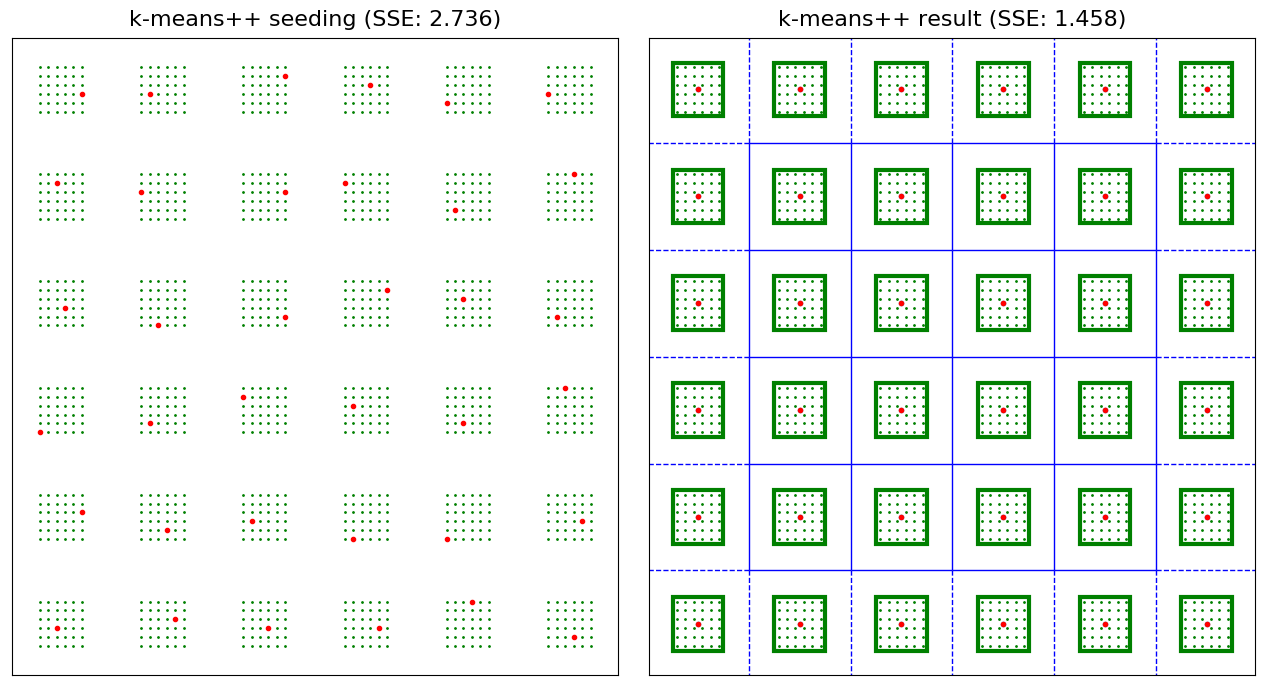}
	\caption[short caption]{\kmp{} optimally solves a given clustering problem $A$-1: 6x6 clusters of 6x6 data points each, k=36. All  clusters correctly receive one center (coded green). In 48-50\% of the simulations this optimal solution was found. Equally often solutions similar to \cref{fig:kmppa1} were found with one empty cluster. In such situations it helps to perform several runs of \kmp{} as done per default in \texttt{scikit-learn} (see text)}
	\label{fig:kmppa1opt}
\end{figure}
\begin{figure}
	\centering
	\includegraphics[width=1.0\linewidth]{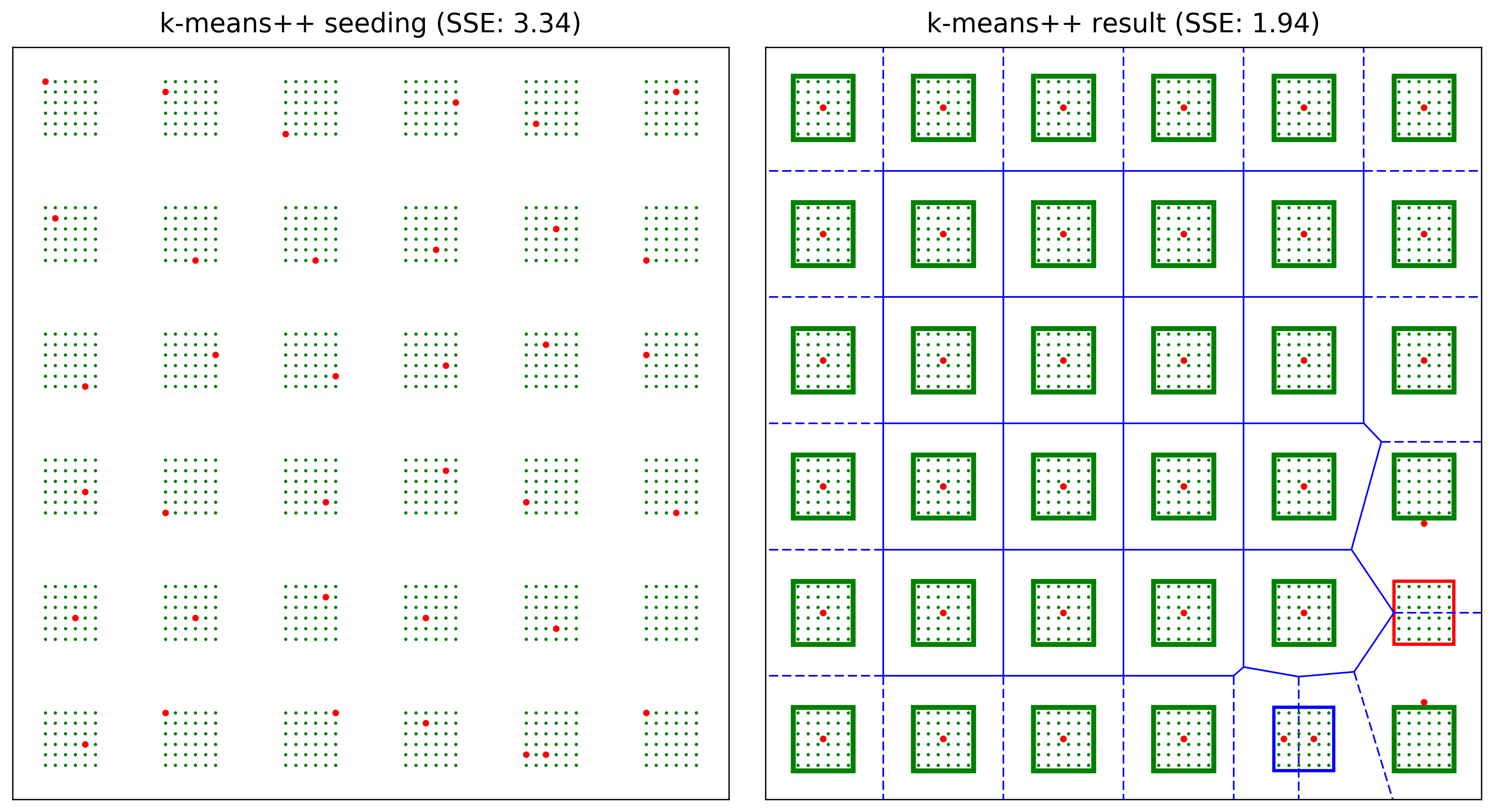}
	\caption[short caption]{\kmp{} near-optimally solves a given clustering problem $A$-1: 6x6 clusters of 6x6 data points each, k=36. 34  clusters correctly receive one center (optimal, coded green), one cluster gets 2 centers (too many, coded in shades of blue) and one cluster gets no center at all (too little, coded red). Such a solution was found in 48-50\% of our simulations. Equally often the optimal solution was found as shown in \cref{fig:kma1_opt}. Looking at this particular simulation one can get an idea why \kmp{} failed to find the optimum here:  In the seeding phase one cluster remained empty (the one colored red on the right-hand figure). Let us denote this cluster as $E$. The reason that $E$ remained empty probably was that the centers placed in the clusters directly above and below $E$ both were positioned quite close to  $E$ leading to relatively small probabilities for the data points in $E$ to be selected as the next center during the \kmp{} seeding phase.}
	\label{fig:kmppa1}
\end{figure}

\section{Why does \kmp{} work so well?}\label{sec:kmpwhy}
Why does \kmp{} work so well? Let us analyze this for the case of a data distribution in 1-D. The calculations are readily generalized to the higher-dimensional case.

\begin{figure}
	\centering
	\includegraphics[width=1.0\linewidth]{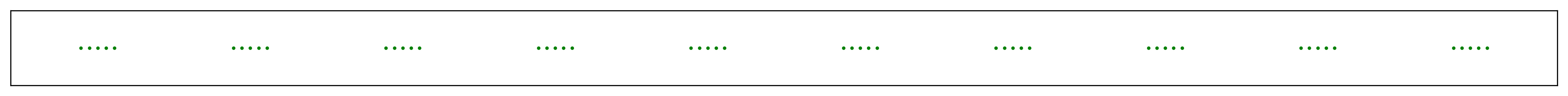}
	\caption[short caption]{One-dimensional data set $A^1$ with $g$ clusters}
	\label{fig:1D}
\end{figure}

Let us consider a data set $A^1$ consisting of $n=g*h$ data point in 1-D space distributed in $g$ separate regions of high density (see \cref{fig:1D}). Each region has a length of $a$ and contains $h=n/g$ points, equally distanced. The distance between neighboring high density regions is $a\eta$. 

Let us first consider a cluster represented by one center which is optimally placed in the center of the cluster. What is the sum of distances in this cluster? If we let $n$ grow towards infinity we can - instead of computing a discrete sum of distances for all points in the cluster - compute the following integral expression:  \begin{equation}F_1 = 2\int_{0}^\frac{a}{2} x^2 \;dx = \frac{a^3}{12}\label{f1}\end{equation}

$F_1$ is proportional to the mean square distance in the cluster and to the cluster width a.

For a cluster not yet covered by a center, all points have distances of at least $a\eta$ to the nearest center. Let us further assume that $\eta$ is so large that we can neglect the difference in distance of points within that cluster and assume all points to have a distance of exactly $a\eta$ to the nearest cluster center. Therefore, the term corresponding to \cref{f1} for the squared distances is 
\begin{equation}F_2 >  2*\int_{0}^\frac{a}{2} (a\eta )^2 \;dx = a^3\eta^2\end{equation}
$F_2$ is proportional to the mean square distance of points in this uncovered cluster to the nearest center and proportional to the cluster width a.

If we assume that $i,\;1 < i<g$ clusters are already covered with one center each, the probability $P_f(i)$ that the next center is placed in one of these "false" clusters can be conservatively estimated as

\begin{equation}P_f(i) < \frac{i \,F_1}{i \,F_1+(g-i)\,F_2} = \frac{i\frac{a^3}{12} }{i\frac{a^3}{12}+(g-i)a^3\eta^2}=\frac{1}{1+12(g/i-1)\eta^2}=c\frac{1}{\eta^2}\end{equation}

Therefore, for any fixed $i$ the following holds:
$$\lim\limits_{\eta \rightarrow \infty}{P_f(i)} = 0$$

In other words,  if the ratio $\eta$ of inter-cluster distance and intra-cluster distance keeps growing the probability of a "wrong" seeding vanishes. Accordingly \kmp{} will almost always place the first $k$ centers optimally in the assumed scenario (data set $\cal{X}$ consisting of $k$ well-separated clusters of equal size). 

The above explains - for our data set $A^1$ with a number of well-separated clusters of similar size - the effectiveness of \kmp{} for the case that $k$ is no larger that the number of clusters in $A$. A similar argument can be made for two two-dimensional data sets $A$ (see \cref{fig:kma1}) or higher-dimensional versions of it. In all cases a growing inter-cluster distance leads to dominating positioning probabilities in so far uncovered clusters.

Perhaps surprisingly the situation changes a lot if one tries to position more centers than there are clusters in the given data set which actually is a typical scenario in applied data analytics where the number of true clusters is usually not known in advance.

\section{Clustering problems which are hard for \kmp}\label{sec:kmphard}

\begin{figure}
	\centering
	\includegraphics[width=0.99\linewidth]{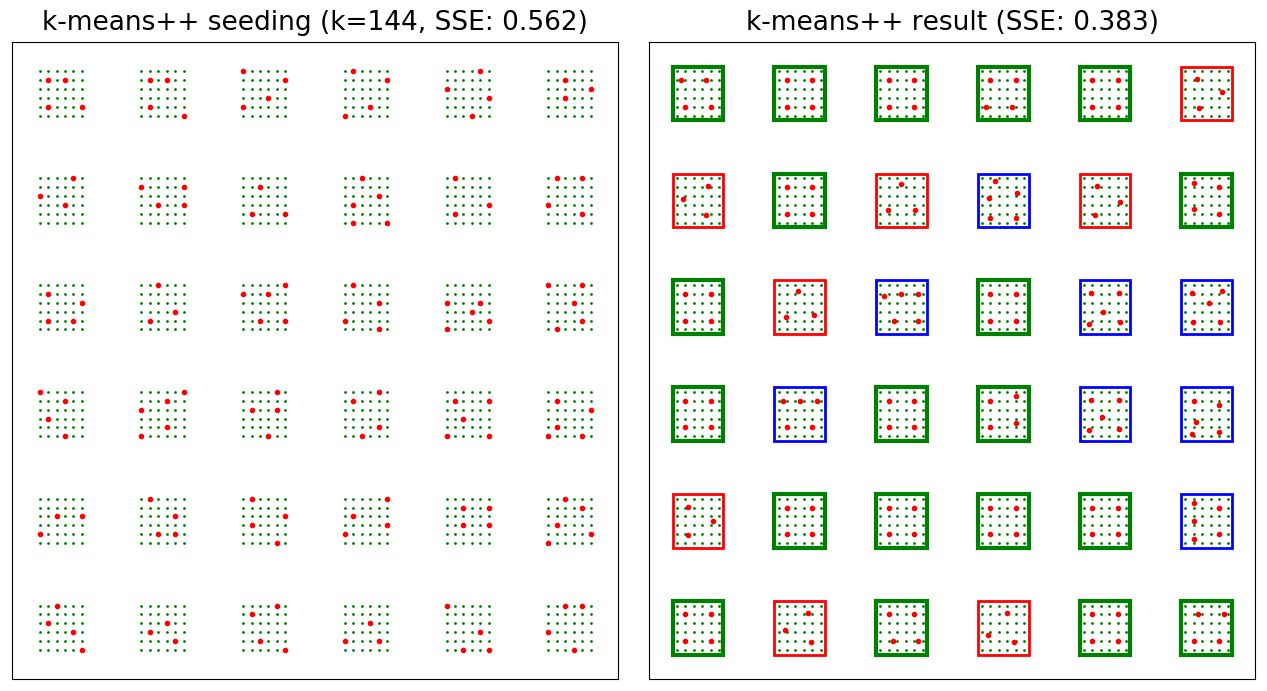}
	\caption[short caption]{\kmp{} approximately solves clustering problem $A$-4: the data is the same as for A-1 (see \cref{fig:kma1,fig:kmppa1}), but in this case k=36*4=144 is chosen. The optimal number of centers per cluster is 4 for this problem. Left: Initialization found by \kmp. 14 of 36 clusters do not contain the optimal number of centers. Right: Final result. The centers have been locally re-distributed by \km, but the number of centers in each particular cluster has not changed. }
	\label{fig:kmppa4}
\end{figure}

 Consider again the data set $A$ from \cref{fig:kma1_opt} with 36 high-density regions. The corresponding clustering problem $A$-1 with $k=36$ has been well solved by \kmp. Let us now consider the problem $A$-4 where the task is to distribute $k=144$ centers over $A$ (i.e.~4 times as many as there are clusters in $A$). The  structure of this data set makes it obvious that for the optimal solution one should place exactly four centers in each cluster. In  \cref{fig:kmppa4} you find a typical result from \kmp.

It can be seen that for many clusters the number of centers placed in them differs from the optimal value (4 in this case). Why is it that \kmp{} can reliably place centers for problem $A$-1 but not so well for problem $A$-4?

As defined in \cref{sec:kmp} \kmp{} uses the (normalized) distance of a data point $x$ from existing centers as probability that $x$ will be the next center. If the number $g$ of clusters is larger than the number $n$ of centers placed so far, this ensures that data points in uncovered clusters have a much larger probability of getting chosen due to their large squared distance to the nearest center. However, if $k > g$ then \kmp{} initially positions one center in each cluster and suddenly the situation changes. Now every data point has a center in its relative vicinity. If further centers are placed they necessarily end up in a cluster already covered by one (later possibly several) centers. Accordingly, they reduce the local error only moderately. This leads to a much higher chance that \kmp{} positions a center sub-optimally and usually leads to improvable results (compared to the optimal distribution of centers, which is unknown in most cases).

In the following we analyze this behavior for the case of the one-dimensional signal distribution $A^1$ already used in \cref{sec:kmpwhy}. This data set consists of $n=g*h$ data point in 1-D space distributed in $g$ separate regions of high density (see \cref{fig:1D}). Each region has a length of $a$ and contains $h=n/g$ points, equally distanced. The distance between neighboring high density regions is $a\eta$. We like to investigate the problem to place $2*g$ centers such that the SSE is minimized. It is easy to see that the optimal solution for this problem consists of 2 centers per cluster placed in the centroid of the first and second half of each cluster. How likely is it that \kmp{} does find this configuration?

If we assume $\eta$ to be large we can - according to the analysis in \cref{sec:kmpwhy} assume that each of the first $g$ centers will be placed in a different cluster. The next center will necessarily be  placed in one of the existing clusters.  This cluster then has two centers which is correct at that point. How likely however is it, that also the remaining $g-1$ centers will be placed correctly, i.e. such that no cluster contains more than 2 centers? 

For simplicity we will assume that in a cluster with one center this center is placed  in the centroid of the cluster and that in a cluster with 2 centers these centers are already placed to minimize the SSE in this cluster, i.e. each in the centroid of one half of the cluster. As in \cref{sec:kmpwhy} we will consider the continuous case by interpreting each cluster as finite segment of length $a$ and we use integrals instead of discrete sums to compare error values for different configurations.

\begin{figure}
	\centering
	\includegraphics[width=1.0\linewidth]{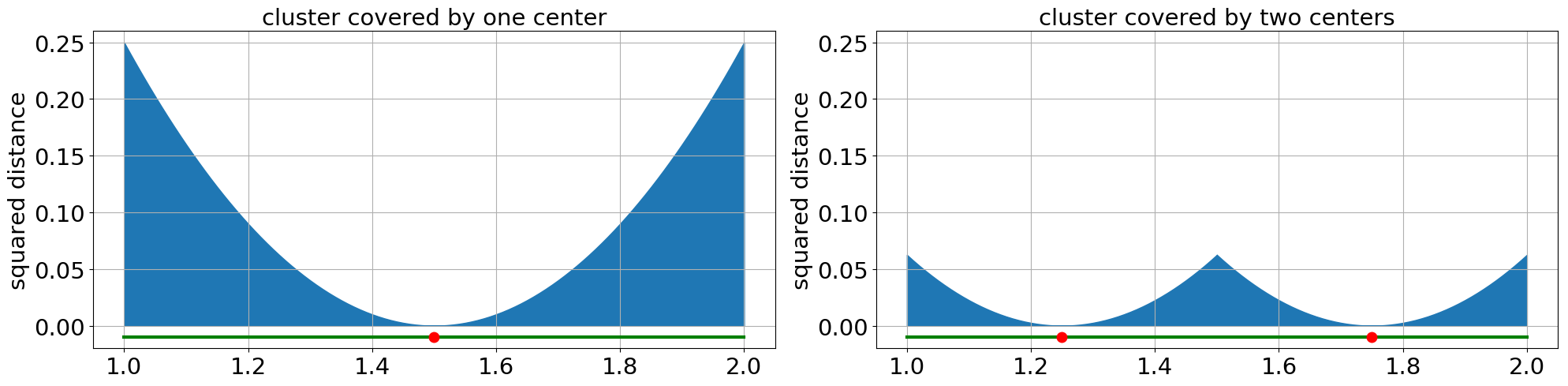}
	\caption[short caption]{ Squared distance within a cluster of length $a=1$. Left: cluster covered by one center. Right: cluster covered by two centers }
	\label{fig:clus12}
\end{figure}

Let us first compute the integral $F_1$ of the squared distance for a cluster of length $a$ covered by one single center positioned in the middle of the cluster (see \cref{fig:clus12}, left side):

\begin{equation}F_1 = 2\int_{0}^{\frac{a}{2}}x^2 dx  = 2 \left [ \frac{1}{3}x^3 \right]_{0}^{\frac{a}{2}}=2*\frac{a^3}{24}=\frac{a^3}{12}=c\;\;\;\mbox{(for some $c$)}\end{equation}
If a cluster is covered by two centers we assume that they are optimally positioned at 25\% and 75\% of its length (see \cref{fig:clus12}, right side). The corresponding integral $F_2$ is the following:
\begin{equation}F_2 = 4\int_{0}^{\frac{a}{4}}x^2 dx  = 4 \left [ \frac{1}{3}x^3 \right]_{0}^{\frac{a}{4}}=4*\frac{a^3}{192}=\frac{a^3}{48}=\frac{c}{4}\end{equation}

 As defined in \cref{sec:kmp} \kmp{} uses the distances as probabilities for placing further centers so $F_1$ is proportional to the probability that a new center is placed in a particular cluster with one center and $F_2$ is proportional to the probability that a new center is placed in a particular cluster with two centers. Since for the following only the relative sizes of $F_1$ and $F_2$ are needed, we can replace $F_1$ by $c$ and $F_2$ by $\frac{c}{4}$.

Let us assume that $g+i$ centers have been placed correctly by \kmp{} among the $g$ clusters (this means that $i$ clusters now have 2 centers). How probable is it that the next center will be placed correctly as well, i.e. in a cluster having only one center so far? To compute this we have to compare the probabilities of the $(g-i)$ "correct" cases to those of all cases:

\begin{equation}
P_{corr}(i,g)=\frac{c*(g-i)}{c*(g-i)+\frac{c}{4}*n}=\frac{g-i}{g-\frac{3}{4}i}
\end{equation}

To compute the probability that all of the $2*g$ centers are placed optimally  (assuming $g+1$ centers were placed correctly already) we have to compute the product of  values $P_{corr}(i,g)$ for $i\in \{1,...,\,g-1\}$:

\begin{equation}
P_{corr}(g)=\prod_{i=1}^{g-1}P_{corr}(i,g)=\prod_{i=1}^{g-1} \frac{g-i}{g-\frac{3}{4}i}
\end{equation}
\begin{figure}
	\centering
	\includegraphics[width=0.99\linewidth]{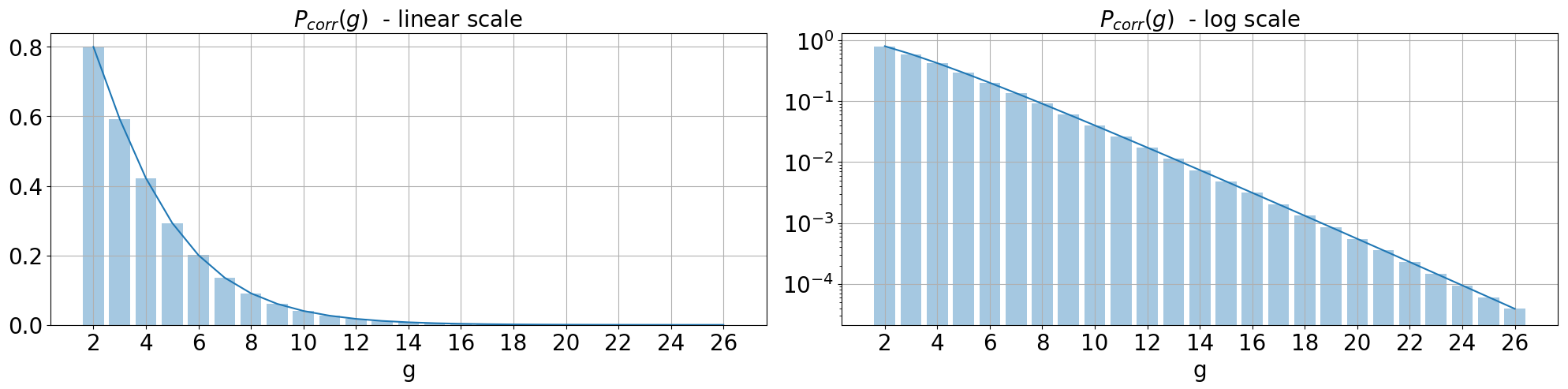}
	\caption[short caption]{Probability $P_{corr}(g)$ that \kmp{} correctly places centers g+1 to 2*g for the clustering problem $A^1$-2 (position $2*g$ centers while the data set is 1-D and has $g$ clusters as shown in \cref{fig:1D}) }
	\label{fig:xprob}
\end{figure}

$P_{corr}(g)$ decays exponentially quickly with $g$. In \cref{fig:xprob} the values of $P_{corr}(g)$ are graphically displayed and already $P_{corr}(24)$ is less than $10^{-4}$. This means that w.r.t. data set $A^1$  - a one-dimensional data set with $g$ clusters - \kmp{} is very unlikely to position the $2*g$ centers such that the optimal configuration with 2 centers per clusters is achieved. This is the case for all values of $g$ which are not trivially small. 

This result is in sharp contrast to the result for the seemingly similar problem to distribute exactly $g$ centers for data sets like $A^1$. In this case by choosing a large distance $\eta a$ among the clusters it can be made arbitrarily probable that \kmp{} finds the optimal configuration (one center in each cluster).

Once there is a center in each cluster, however, the inter-cluster distance is rendered meaningless. A further center in any cluster reduces the local error only moderately (e.g. by a factor of 4 if we add a second center to a cluster having one center so far, see \cref{fig:clus12}). This again means that if we already have many clusters with two centers, it becomes very probable that a third center is placed in one of them by \kmp{} leading to a sub-optimal configuration.

The analyzed behavior for 1-D could be extended to other data distribution and to higher data dimensions. Completely general statements however are probably difficult to establish since there is a large dependency on the particular structure of the given data set which for real data sets is usually not known. Therefore we will rather concentrate on presenting a method to improve upon the results of \kmp{} and demonstrate its effectiveness by systematic comparative simulations with data sets of varying size and dimensionality. 

\section{\kmu}\label{sec:kmu}
How can the results obtained with \kmp{} be further improved? \kmp{} employs a careful seeding step followed by standard \km{} and thus ends in a (usually local) minimum of the error function $\phi(\mathcal{C},\mathcal{X})$ from \cref{eq:1}. One idea to improve such a solution is to move single centers to other positions and afterwards let \km{} find a - hopefully better - local minimum. To make this computationally efficient and guarantee convergence, however, one needs to carefully select both the centers to be moved and their respective target positions. Also a stopping criterion needs to be defined.

Which center should we move? Following the approach proposed by \cite{Fritzke97a} we investigate how "useful" each center is for error reduction. This can be quantified by removing this center and comparing the resulting error with the current error. Thus we define the \emph{utility} $U(c_i)$ of a center $c_i$ as

\begin{equation}
U(c_i) = \phi(\mathcal{C}\setminus \{c_i\},\mathcal{X}) - \phi(\mathcal{C},\mathcal{X})\label{eqnutility}
\end{equation}

\begin{figure}[bh]
	\centering
	\includegraphics[width=0.60\linewidth]{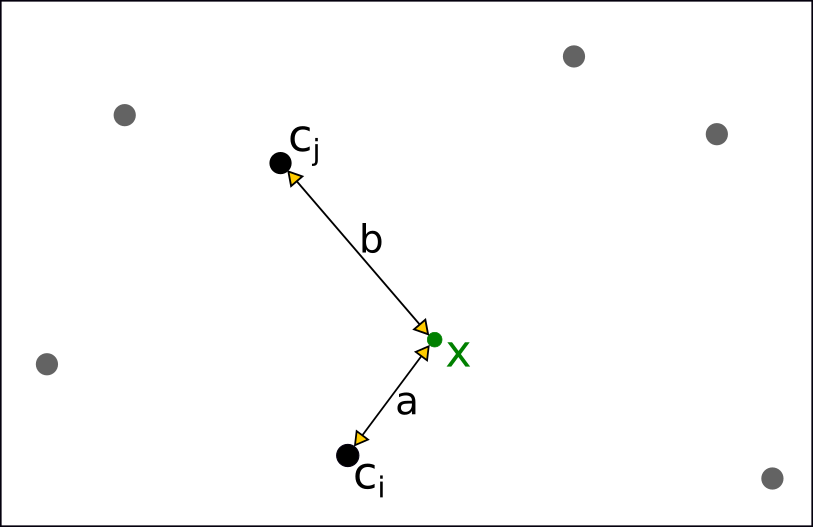}
	\caption[short caption]{Utility illustrated and alternatively defined: Let center $c_i$ be the closest center for a data point $x$ with $a=\|x-c_i\|$ and let center $c_j$ be the second-closest center to $x$ with $b=\|x-c_j\|$. Then the data-point-specific utility of center $c_i$ for encoding data point $x$ can be expressed as $U_x(c_i)=b^2-a^2$. This term is the additional squared error which occurred for encoding $x$ if $c_i$ did not exist (thus the term utility).  Accordingly the total utility of $c_i$ is the sum of the data-point-specific utilities over the Voronoi set $\mathcal{C}_i$ (see \cref{eqn:voro}) of all data points for which $c_i$ is the nearest center: $U(c_i)=\sum_{x\in\mathcal{C}_i } U_x(c_i)$. It is easy to see that this definition is equivalent to \cref{eqnutility}.}
	\label{fig:utility}
\end{figure}

The utility of a center is a measure how much we need this center or how easy its "task" of reducing error can be overtaken by neighboring centers. For example in the pathological case where two centers have the same position they both have a utility of zero since one could remove either one of them without increasing the error: the remaining center would cover the associated data points as well as the two centers did before. This case will not occur with \km{} but all other things being equal one can expect the utility of centers with close neighbors being lower than those of more isolated centers. For an alternative but equivalent definition of the utility see \cref{fig:utility}.

We now define the "least useful" center $\lambda$ as follows:

\begin{equation}
\lambda = \argmin\limits_{c_i\in\mathcal{C}}\;U(c_i) = \argmin\limits_{c_i\in\mathcal{C}}\; \phi(\mathcal{C}\setminus \{c_i\},\mathcal{X}) = \argmin\limits_{c_i\in\mathcal{C}}\; \sum_{x\in\mathcal{X}} \min\limits_{c_j\in\mathcal{C}\setminus \{c_i\}} ||x-c_j||^2 \label{eqnlambda}
\end{equation}

Where should we move the center $\lambda$? Since our goal is to reduce the overall error function one rather straightforward approach is to move $\lambda$ to the vicinity of that center $\mu$ currently having the maximal error sum for its Voronoi set.

%

Recall that in \cref{sec:km}, \cref{eqn:voro}  we defined for each $i \in \{1, ...,k \}$ the Voronoi set $C_i$ to be the set of all points in $\mathcal{X}$ that are closer to $c_i$ than they are to $c_j$ for all $j \ne i$.
We now define $\mu$ as the center $c_i$ having the largest summed squared distance over its Voronoi set $C_i$:

\begin{equation}
\mu = \argmax\limits_{c_i\in\mathcal{C}} \sum_{x\in C_i} ||x-c_i||^2\label{eqnmu}
\end{equation}
It is advisable to place $\lambda$ not exactly at the position of $\mu$ since that would cause all points in $C_\mu$ to have identical distances to $\mu$ and $\lambda$. We therefore place center $\lambda$ at the position of $\mu$ plus some small random offset. A simple approach to define what "small" means for a given center $\mu$ is to consider the mean distance of the data points in $\mu$'s Voronoi set $C_\mu$.  
We therefore define $d_\mu$ as
\begin{equation}\label{eqn:muerr}
d_\mu= \sqrt{\frac{1}{|C_\mu|}\sum_{x\in C_\mu} \|x-\mu\|^2 }
\end{equation}
$d_\mu$ gives an indication of the spatial extension of $\mu$'s  Voronoi set. By choosing a small fraction of $d_\mu$, e.g. $\epsilon d_\mu$ with $\epsilon=0.01$, as the length of our offset vector we can be confident  that the new position of $\lambda$ will be "near" $\mu$. 

Having a length, we still have to choose a direction for our offset vector. A principled and informed choice would be the direction of largest variance in the Voronoi set of $\mu$, i.e. the unit eigenvector corresponding to the largest eigenvalue of the covariance matrix of $C_\mu$. 

Instead - both to have some non-determinism and to save the eigenvector computation in each step - we simply use a random vector from the $d$-dimensional unit hypersphere and rely on the following \km{} phase to find good configurations. Choosing such a vector with uniform probability density is not completely trivial, however. For example normalizing a random vector from the $d$-dimensional hypercube would cause probability peaks in the directions of the corners of this hypercube. But it has been shown \citep{Marsaglia1972} that a uniformly distributed unit random vector from the $d$-dimensional unit hypersphere can be constructed as follows:

\begin{enumerate}
	\item Generate $d$ Gaussian random variables $x_1,\,x_2,\, ...,\, x_d$
	\item Return the vector
	\begin{equation}
	\xi = \frac{1}{\sqrt{x_1^2,\,x_2^2,\, ...,\, x_d^2}} [x_1,\,x_2,\, ...,\, x_d]^T\end{equation}
\end{enumerate} 

Let $u$ be such a random vector. Then we define our offset vector $o$ as

		\begin{equation}\label{eqn:offset}
		o=\epsilon\, d_\mu\, u 
		\end{equation}

The \kmu{} algorithm which we now define, starts with \kmp. Thereafter repeatedly the centers $\mu$ and $\lambda$ are determined and the least useful center $\lambda$ is moved to the position of the center $\mu$ with maximum error (plus a small random offset $o$ which is also applied to $\mu$ itself, but with opposite sign). After each such move standard \km{} is performed and the resulting error is measured. The algorithm terminates as soon as there is no improvement of the error measure. The complete \kmu{} (\km{} with utility) algorithm is specified in \cref{alg:kmu}.

\begin{algorithm}[ht]
	\SetKwFor{Loop}{Loop}{}{EndLoop}
	
	\setstretch{1.25}
	\SetKw{ini}{Seeding:}
	\ini $\mathcal{C}\leftarrow$ (result of \kmp)\;
	$\phi_{\mathrm{best}} \leftarrow \phi(\mathcal{C},\mathcal{X})$ \tcc*[r]{store lowest error so far (from \kmp)}
	$\mathcal{C}_{\mathrm{best}} \leftarrow \mathcal{C}$ \tcc*[r]{store best $\mathcal{C}$ so far (from \kmp)}
	\Loop{}{
		$\lambda \leftarrow \argmin\limits_{c_i\in\mathcal{C}}
		\phi(\mathcal{C}\setminus\{c_i\},\,\mathcal{X})$
		\tcc*[r]{find least useful center}		
		$
		\mu \leftarrow \argmax\limits_{c_i\in\mathcal{C}} \sum\limits_{x\in C_i} ||x-c_i||^2
		$
		\tcc*[r]{find center with max.~local error}
		$u \leftarrow \mbox{(random vector from $d$-dimensional unit hypersphere)}$\;
		$
		d_\mu \leftarrow \sqrt{\frac{1}{|C_\mu|}\sum\limits_{x\in C_\mu} \|x-\mu\|^2 }
		$
		\tcc*[r]{mean distance around $\mu$}	
		$o\leftarrow\epsilon\, d_\mu\, u$
		\tcc*[r]{offset vector, $\epsilon = 0.01$}	
		$\lambda \leftarrow (\mu + o)$
		\tcc*[r]{position $\lambda$ near $\mu$}	
		$\mu \leftarrow (\mu - o)$
		\tcc*[r]{position $\mu$ opposite to $\lambda$ w.r.t.~old $\mu$ value}
		Perform \km{} using the current $\mathcal{C}$ as initial set of centers\;
		\eIf{$\phi(\mathcal{C},\mathcal{X}) < \phi_\mathrm{best}$}{
			$\phi_{\mathrm{best}} \leftarrow \phi(\mathcal{C},\mathcal{X})$ \tcc*[r]{store new lowest error}
			$\mathcal{C}_{\mathrm{best}} \leftarrow \mathcal{C}$ \tcc*[r]{store new best $\mathcal{C}$}
		}{
			break
			\tcc*[r]{exit loop}	
		}
	}
	\Return{$\mathcal{C}_{\mathrm{best}} $}\;
	
	\caption{The \kmu{} algorithm}
	\label{alg:kmu}
\end{algorithm}

\begin{figure}
	\centering
	\includegraphics[width=0.99\linewidth]{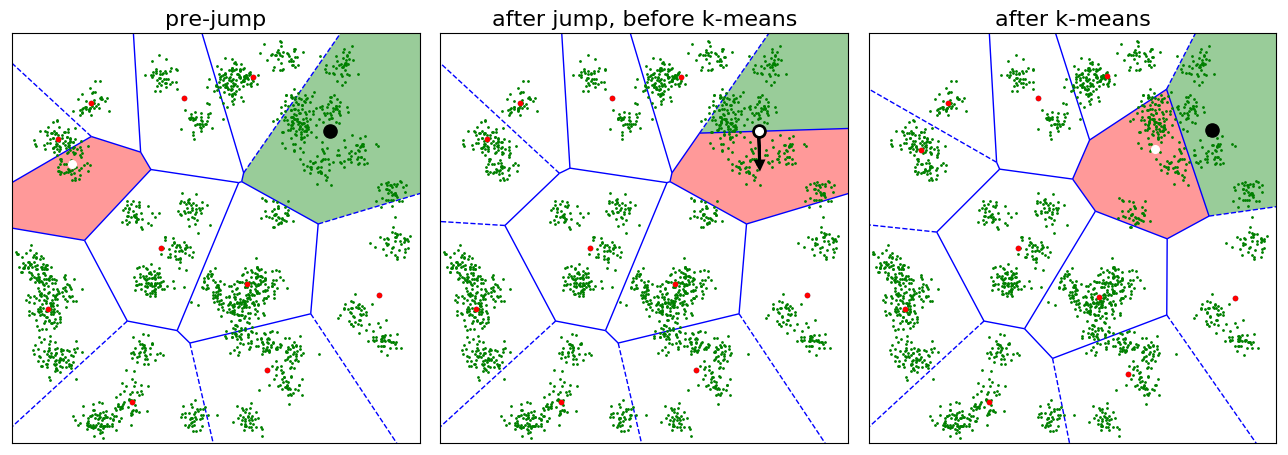}
	\caption[short caption]{Detail view of a jump. The left figure shows a configuration where the initial run of \kmp{} (the first phase of \kmu) has converged. The center $\mu$ with largest local summed squared error (SSE) is shown black (with a green Voronoi region). The center  $\lambda$ with the lowest utility is shown in white (with a reddish Voronoi region). The middle figure shows the configuration directly after performing a jump from $\lambda$ to $\mu$ but before applying \km. A small random offset vector has been applied to both $\mu$ and $\lambda$ but in opposite direction. The direction of the offset vector is shown as an arrow. The offset vector is a normal vector of the $(d-1)$-dimensional hyperplane (line for 2-D data) which divides the previous Voronoi region of $\mu$ into the new Voronoi regions of $\mu$ (green) and $\lambda$ (reddish). The right figure shows the result of the \km{} run directly following the jump. $\mu$ and $\lambda$ are now separated again each being in the center of gravity of their respective Voronoi set.}
	\label{fig:closeup0}
\end{figure}

What does locally happen during a jump in \kmu? An example is shown in \cref{fig:closeup0}. At the previous position of $\lambda$ the data points so far belonging to $\lambda$'s Voronoi set are now partitioned among the neighboring centers whose Voronoi regions are correspondingly enlarged. Assuming that $o$ is very small the centers  $\mu$ and $\lambda$ now both have positions very close to the previous position of $\mu$ but offset in opposite directions. The previous Voronoi set of $\mu$ is therefore partitioned into two subsets $C_\mu$ and $C_\lambda$ divided by a $(d-1)$-dimensional hyperplane. This hyper plane  contains the previous position of $\mu$ and is also a normal plane to the offset vector $o$. Since $o$ is created from a random vector, the orientation of this hyperplane and the resulting partitioning  is random as well. This observation will become relevant in \cref{sec:kms}.

\begin{figure}
	\centering
	\includegraphics[width=0.99\linewidth]{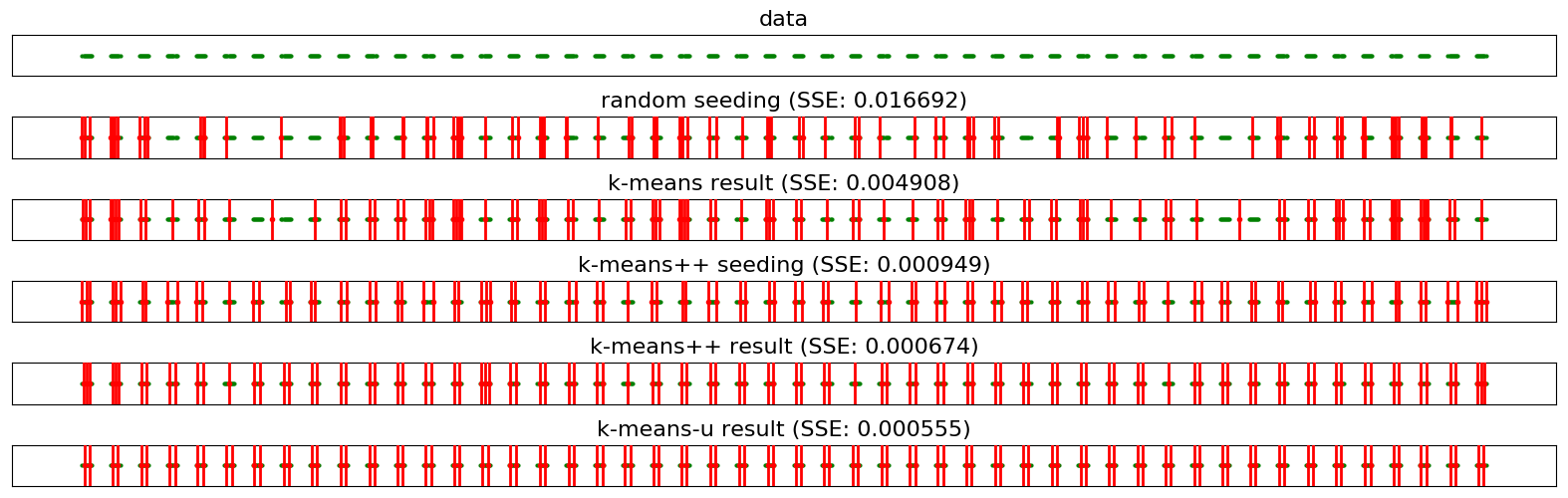}
	\caption[short caption]{Results of \km, \kmp{} and \kmu{} for a one-dimensional distribution consisting of 50 clusters. The task is to distribute 100 centers among the data such that the SSE is minimized. \km{} suffers from the  poor (but typical) random seeding. The \kmp{} seeding is able to distribute the centers much better but still makes some "errors" wrt.~the given data set (in accordance with the analysis from \cref{sec:kmphard}).  These errors (positioning either one or three centers in a cluster) cannot be corrected by the following \km{} phase. \kmu{}, starting from the displayed result of \kmp{}, is able to generate a result with two centers in each cluster, a necessary condition for  the optimum in this case. The error achieved by \kmu{} is 17.7\% lower than that from \kmp. }
	\label{fig:oneD}
\end{figure}

In \cref{fig:oneD} a simple one-dimensional example illustrates typical results of \km, \kmp{} and \kmu. The data set used is a variant of data set $A^1$ which was used in \cref{sec:kmpwhy} and \cref{sec:kmphard} to illustrate easy and hard problems for \kmp.


\begin{figure}
	\centering
	\includegraphics[width=0.99\linewidth]{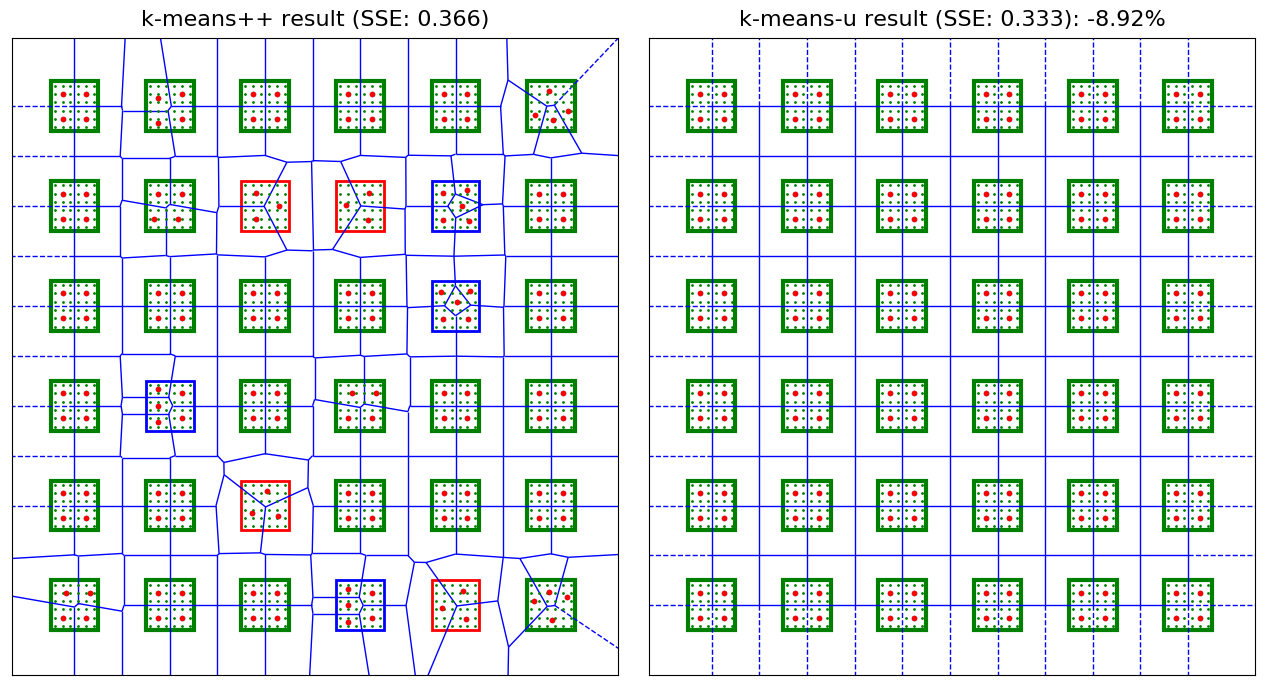}
	\caption[short caption]{\kmu{} optimally solves clustering problem $A$-4,  Left: Result produced by \kmp. 8 of 36 clusters do not contain the optimal number of centers. Right: Result of \kmu. The non-local jumps lead in this case to the optimal solution}
	\label{fig:kmua4}
\end{figure}

\begin{figure}
	\centering
	\includegraphics[width=0.99\linewidth]{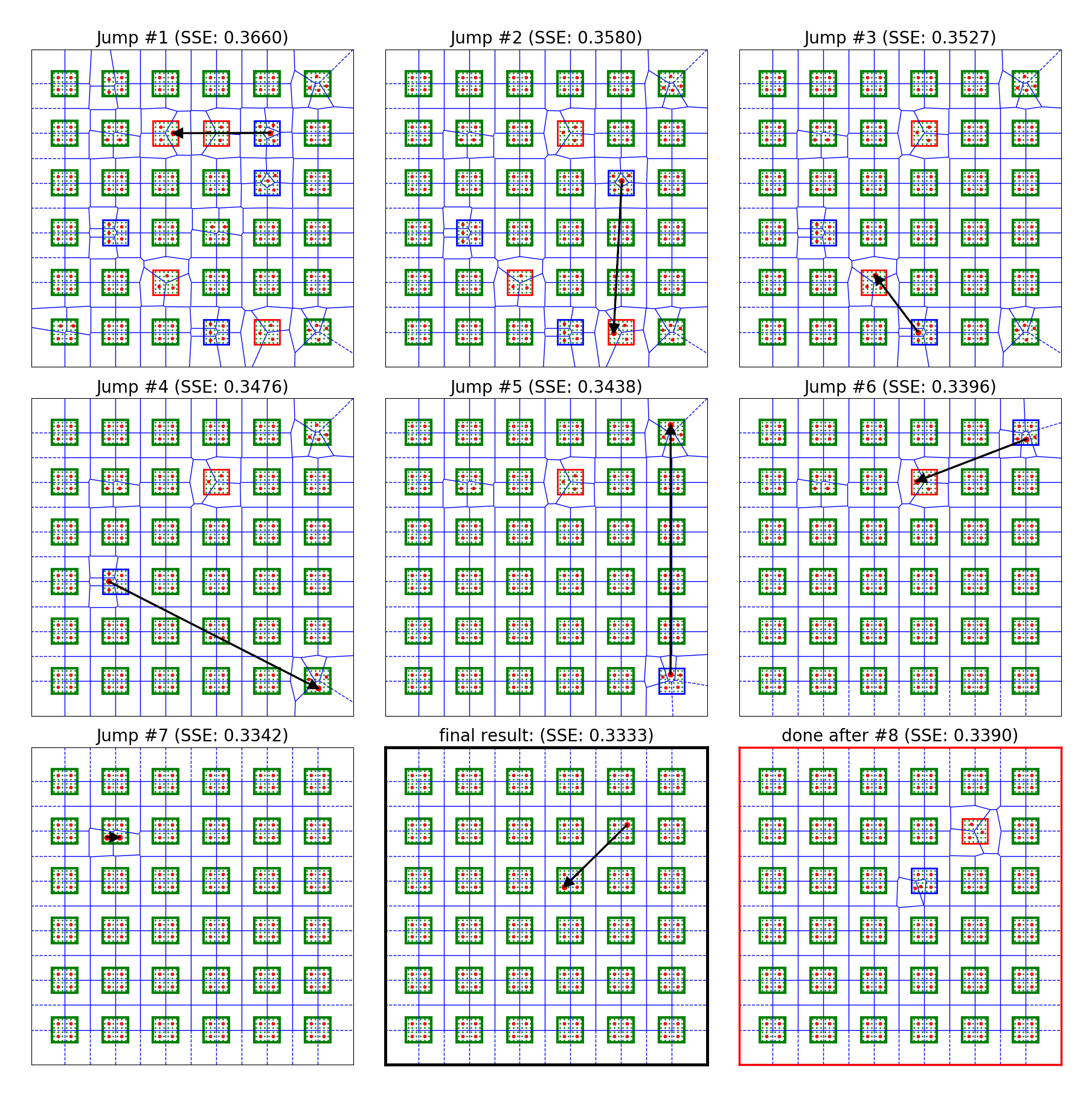}
	\caption[short caption]{\kmu{} performs non-local jumps: starting from the result of \kmp{} the derived sequence of jumps is displayed. Each sub figure shows a converged configuration of \km{}(resp. \kmp{} in the initial figure) and the non-local jump to be performed based on this configuration. The displayed SSE is always "pre-jump". Jumps are performed until the error stops dropping or even raises. At that point the last-but one local minimum is returned as the final result of the algorithm.}
	\label{fig:arrows}
\end{figure}

\begin{figure}
	\centering
	\includegraphics[width=0.99\linewidth]{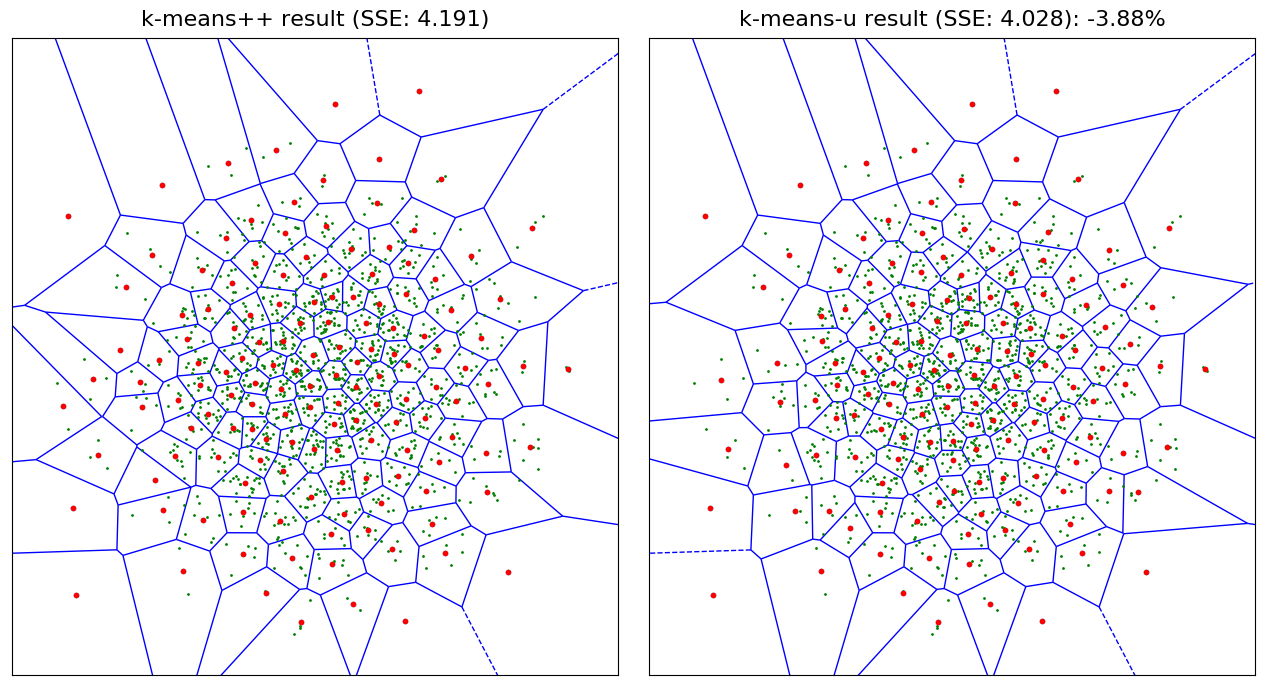}
	\caption[short caption]{\kmp{} and \kmu{} applied to mixture of overlapping Gaussians with 1296 data points and 36 centers,  Left: Result produced by \kmp. Right: Result of \kmu with large relative SSE improvement}
	\label{fig:kmua4GMM}
\end{figure}

\Cref{fig:kmua4} contains a 2-dimensional example where \kmu{} found a considerable improvement over the result of \kmp.
In \cref{fig:arrows} the sequence of non-local jumps leading to the result in \cref{fig:kmua4} is displayed.
\Cref{fig:kmua4GMM} demonstrates that \kmu{} is also able to find improvements for more natural data sets. In this case the data is from a mixture of overlapping Gaussians. The number of data points and the number of centers is exactly as in \cref{fig:kmua4}. The improvement over \kmp{} is nearly 4\% in this example, even though it is hard to see the difference between the two solutions.

Note: In principle also another seeding method than \kmp{} can be used (e.g. standard \km{} or even random seeding) but in simulations \kmp{} led to the best results for the following \kmu{} algorithm.
\clearpage        
\section{An occasional problem of \kmu: too early termination}\label{sec:occasional}
Sometimes we observed in simulations that \kmu{} finished very early, i.e. at a point in time  when there were seemingly many optimization opportunities left. Initially we suspected a programming error, but indeed in all investigated cases the most recent jump had led to a particular poor - but stable - configuration causing a relative increase of the SSE and thus a termination of \kmu{} according to its definition. 

\begin{figure}
	\centering
	\includegraphics[width=0.99\linewidth]{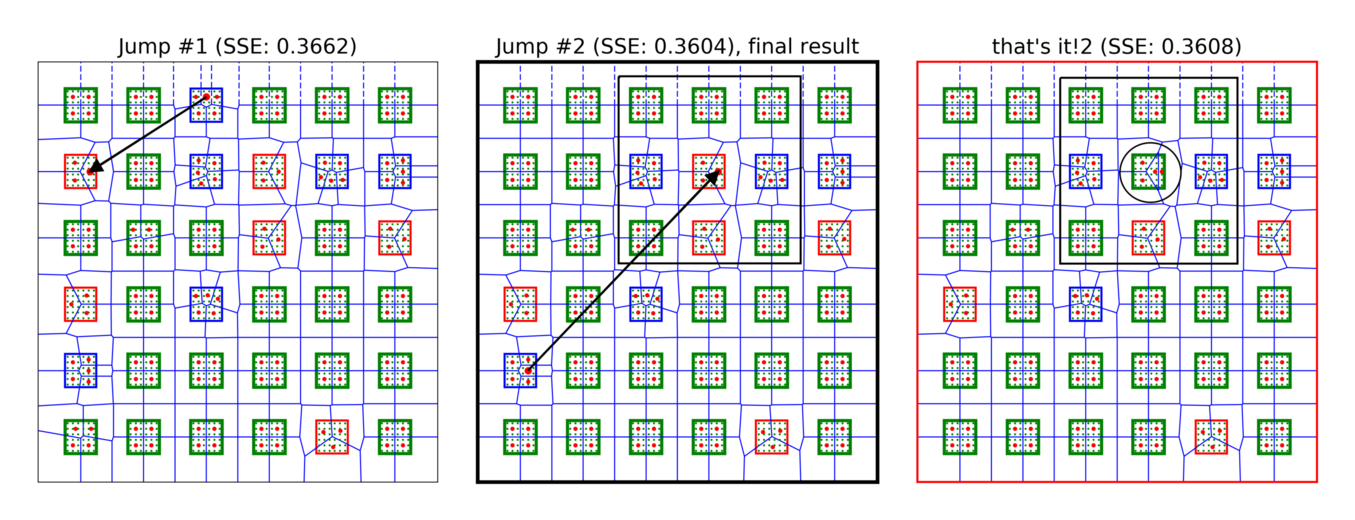}
	\caption[short caption]{Too early stopping of \kmu:  Since already the second jump led to an increase of the SSE (right-most figure), the algorithm returns a result with many non-optimally covered clusters (center figure). The relevant region of the input space is indicated by a box and is shown enlarged in \cref{fig:arrowspoordetail}.}
	\label{fig:arrowspoor}
\end{figure}

\begin{figure}[!tbp]
	\hfill
	\begin{subfigure}[t]{0.47\textwidth}
		~\hfill{\includegraphics[width=0.85\textwidth]{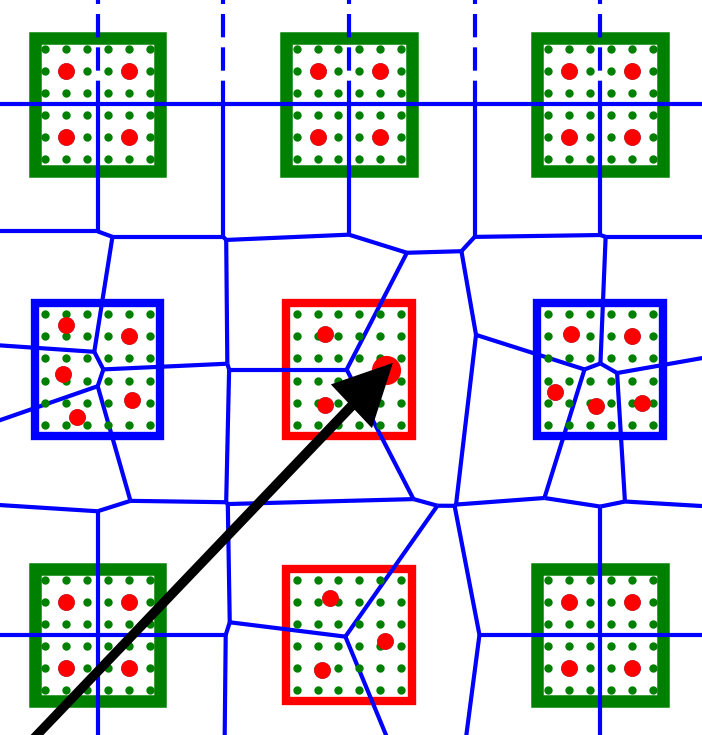}}\hfill~
		\caption{Before jump 2 from \cref{fig:arrowspoor}. Center $\mu$ with largest local error is enlarged.}
	\end{subfigure}
	\hfill
	\begin{subfigure}[t]{0.47\textwidth}
		~\hfill{\includegraphics[width=0.85\textwidth]{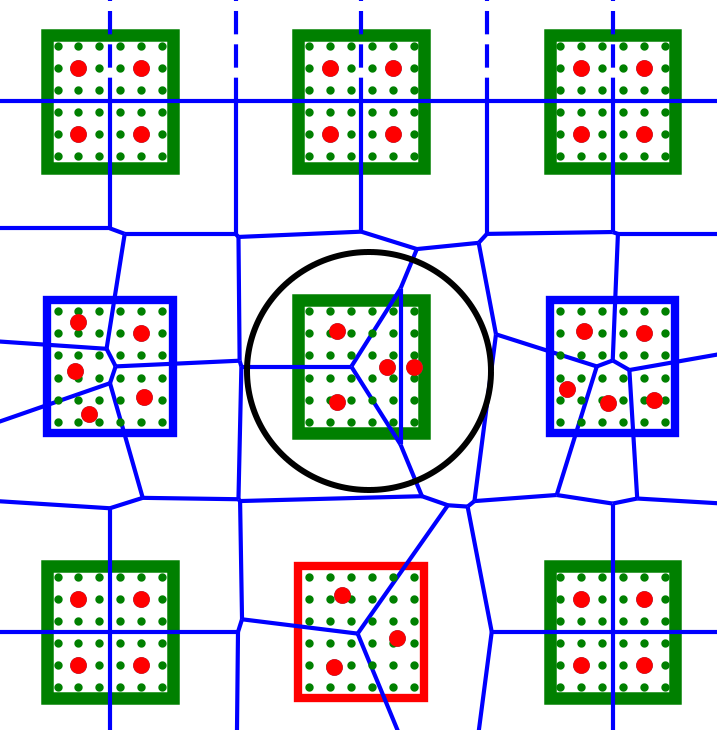}}\hfill~
		\caption{Result of \km{} phase: the encircled cluster has a very high local error}
		\label{fig:f2}
	\end{subfigure}
	\caption{Details of the  simulation sequence from \cref{fig:arrowspoor} where \kmu{} terminated with a quite sub-optimal result}
	\hfill
	\label{fig:arrowspoordetail}
\end{figure}

In \cref{fig:arrowspoor} such a simulation sequence is depicted. \kmu{} already terminated after two jumps because the SSE had increased after performing the second jump. In \cref{fig:arrowspoordetail} a detail view of the cluster causing the error increase is shown. It does contain the optimal number (4) of centers for this problem, but their arrangement determined by \km{} is such that two centers are very close to each other and both have elongated Voronoi regions and therefore a relatively high distance to the member points of their respective Voronoi sets.

Fortunately these poor configurations seem to be  relative rare. Due to the associated high SSE values, however, they have the potential to deteriorate mean performance statistics.   Given that in \kmu{} the re-positioning of the two centers affected from a jump  ($\mu$ and $\lambda$) is based on a random vector, one relatively easily comes up with the idea to re-do the positioning in such cases. This leads to an extension of our original algorithm described in the next section.

\section{\kms}\label{sec:kms}

As exemplified in \cref{sec:occasional} runs of \kmu{} may end too early due to poor local configurations \km{} runs into after a jump. In \cref{sec:kmu} it was discussed what happens at a jump. Let us analyze this here in more detail: Directly after the repositioning the centers $\mu$ and $\lambda$ divide\footnotemark[2] the Voronoi set $C_\mu$ (see \cref{eqn:voro}) of $\mu$ among them using a $(d-1)$-dimensional hyperplane going through the previous position of $\mu$ and having the offset vector $o$ (\cref{eqn:offset}) as its normal vector.\footnotetext[2]{Strictly speaking this is true only in the limiting case when the length of  $o$ goes to zero, since with a non-vanishing vector $o$ there may be some data points previously associated with other centers for which now either $\mu$ or $\lambda$ is the closest center. If $o$ goes to zero however, the combination of the Voronoi regions of $\mu$ or $\lambda$ approaches the old Voronoi region of $\mu$ with arbitrary precision.}

The orientation of $o$ depends on a random vector drawn from a $d$-dimensional hypersphere and determines how the data points previously associated with $\mu$ are distributed between $\mu$ and $\lambda$. 
\begin{figure}
	\centering
	\includegraphics[width=0.99\linewidth]{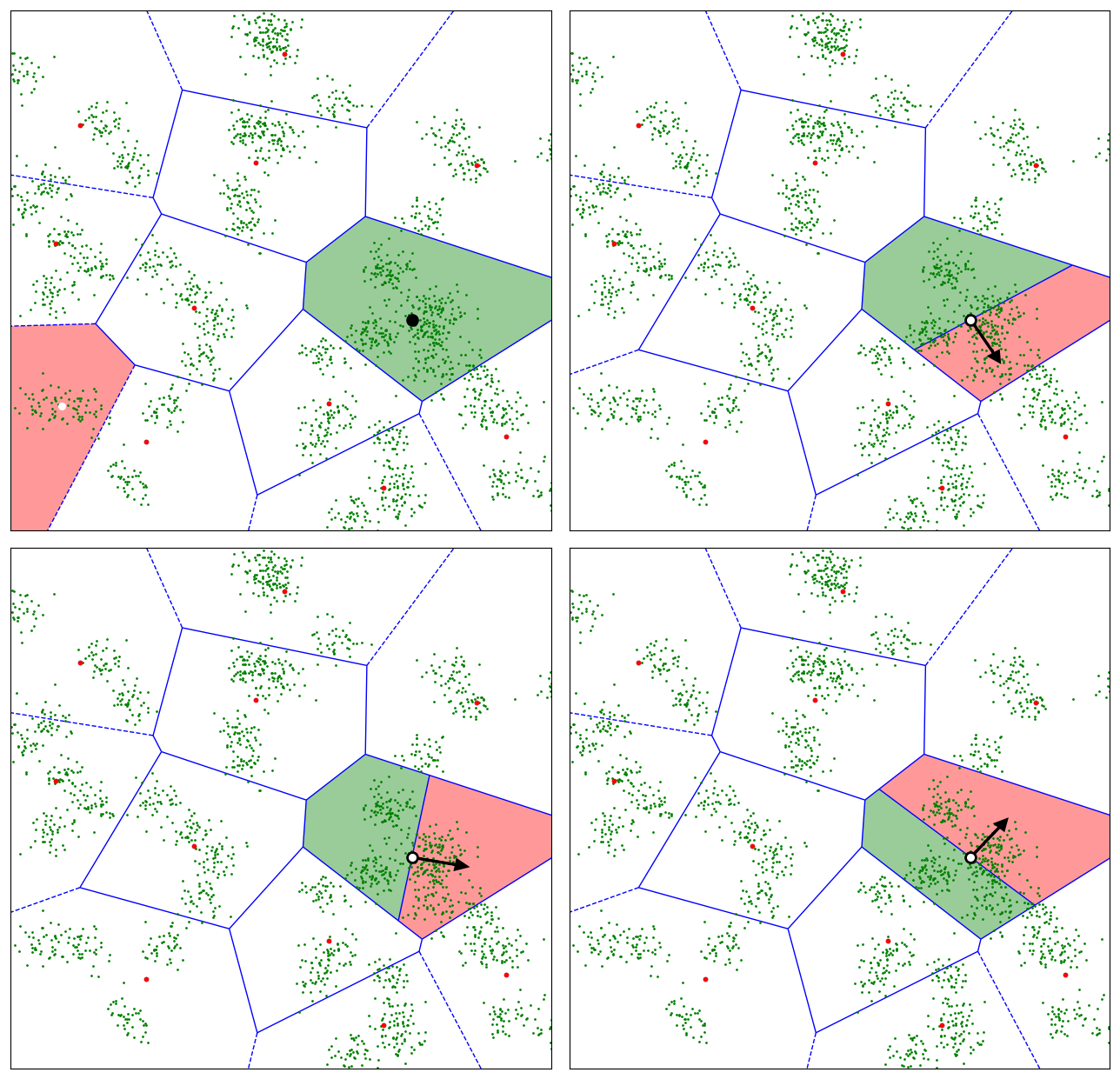}
	\caption[short caption]{Different results of a jump depending on the orientation of the offset vector. The upper left figure shows a configuration where \km{} has converged. The center $\mu$ with the largest local summed squared error (SSE) is shown black (with a green Voronoi region). The center $\lambda$ with the lowest utility is shown in white (with a reddish Voronoi region). The other three images show three different configurations obtained by performing a "jump" from $\lambda$ to $\mu$ and applying a small random offset vector to both $\mu$ and $\lambda$ but in opposite directions. The direction of the offset vector is shown as an arrow in each case. The offset vector is a normal vector of the $(d-1)$-dimensional hyperplane (line for 2-D data) which divides the previous Voronoi region of $\mu$ into the new Voronoi regions of $\lambda$ (reddish) and  $\mu$ (green).  }
	\label{fig:closeup}
\end{figure}

Each random choice leads with probability one (choosing two collinear vectors from a continuous distribution on the hypersphere has probability zero) to a different orientation of the hyperplane and likely to a different partitioning of the affected data points (see \cref{fig:closeup} for an example). The \km{} phase following every jump leads to results depending on these partitionings. Different partitionings likely lead to different results of \km.

Based on the above observations we propose the following simple extension of the \kmu{} algorithm which we call \kms{} (see \cref{alg:kms}). Instead of immediately terminating after an error increase we allow a small finite number $retry_{max}$ of retries of the most recent jump. Due to the random choice of the offset vector these retries possibly end up in a configuration with lower error and allow a continuation of \kmu{}, sometimes for many steps. Once a retry was successful we "reset" the retry counter so the specified number of retries is again available at a later stage which will be at a lower error level than the previous retry sequence (since we just improved our "best solution"). This leads to a strictly monotone sequence of error values of the respective best solution after every retry sequence until the algorithm terminates. Since we never try to improve a configuration with a higher error than the best solution found so far, this retry procedure can be interpreted as randomized "greedy search".

\begin{algorithm}[ht]
	\SetKwFor{Loop}{Loop}{}{EndLoop}
	
	\setstretch{1.25}
	\SetKw{ini}{Seeding:}
	\ini $\mathcal{C}\leftarrow$ (result of \kmp)\;
	$\phi_{\mathrm{best}} \leftarrow \phi(\mathcal{C},\mathcal{X})$ \tcc*[r]{store lowest error so far (from \kmp)}
	$\mathcal{C}_{\mathrm{best}} \leftarrow \mathcal{C}$ \tcc*[r]{store best $\mathcal{C}$ so far (from \kmp)}
	\textcolor{red}{
		$retry_{\mathrm{max}} \leftarrow n$
		\tcc*[r]{$n \in \{0,\,1,\,2,\,...\}$}	}
	\textcolor{red}{$retry \leftarrow 0$
		\tcc*[r]{initialize retry counter}}		
	
	\textcolor{red}{
		\Repeat{$retry > retry_\mathrm{max}$}{
			\textcolor{black}{	
				\Loop{}{
					$\lambda \leftarrow \argmin\limits_{c_i\in\mathcal{C}}
					\phi(\mathcal{C}\setminus\{c_i\},\,\mathcal{X})$
					\tcc*[r]{find least useful center}		
					$
					\mu \leftarrow \argmax\limits_{c_i\in\mathcal{C}} \sum\limits_{x\in C_i} ||x-c_i||^2
					$
					\tcc*[r]{find center with max.~local error}
					$u \leftarrow \mbox{(random vector from $d$-dimensional unit hypersphere)}$\;
					$
					d_\mu \leftarrow \sqrt{\frac{1}{|C_\mu|}\sum\limits_{x\in C_\mu} \|x-\mu\|^2 }
					$
					\tcc*[r]{mean distance around $\mu$}	
					$o\leftarrow\epsilon\, d_\mu\, u$
					\tcc*[r]{offset vector, $\epsilon = 0.01$}	
					$\lambda \leftarrow (\mu + o)$
					\tcc*[r]{position $\lambda$ near $\mu$}	
					$\mu \leftarrow (\mu - o)$
					\tcc*[r]{position $\mu$ opposite to $\lambda$ w.r.t.~old $\mu$ value}
					Perform \km{} using the current $\mathcal{C}$ as initial set of centers\;
					\eIf{$\phi(\mathcal{C},\mathcal{X}) < \phi_\mathrm{best}$}{
						$\phi_{\mathrm{best}} \leftarrow \phi(\mathcal{C},\mathcal{X})$ \tcc*[r]{store new lowest error}
						$\mathcal{C}_{\mathrm{best}} \leftarrow \mathcal{C}$ \tcc*[r]{store new best $\mathcal{C}$}
						\textcolor{red}{$retry \leftarrow 0$
							\						\tcc*[r]{improvement! reset retry counter}}		
					}{
						break
						\tcc*[r]{exit loop}	
					}
				}
			}
			$retry \leftarrow retry + 1$\;
			$\mathcal{C} \leftarrow \mathcal{C}_{\mathrm{best}}$ \tcc*[r]{rewind to best solution so far (=previous)}
		}
	}
	\Return{$\mathcal{C}_{\mathrm{best}} $}\;
	
	\caption{The \kms{} algorithm. Additions to \kmu{} shown \textcolor{red}{red}}
	\label{alg:kms}
\end{algorithm}

\begin{figure}
	\centering
	\includegraphics[width=0.95\linewidth]{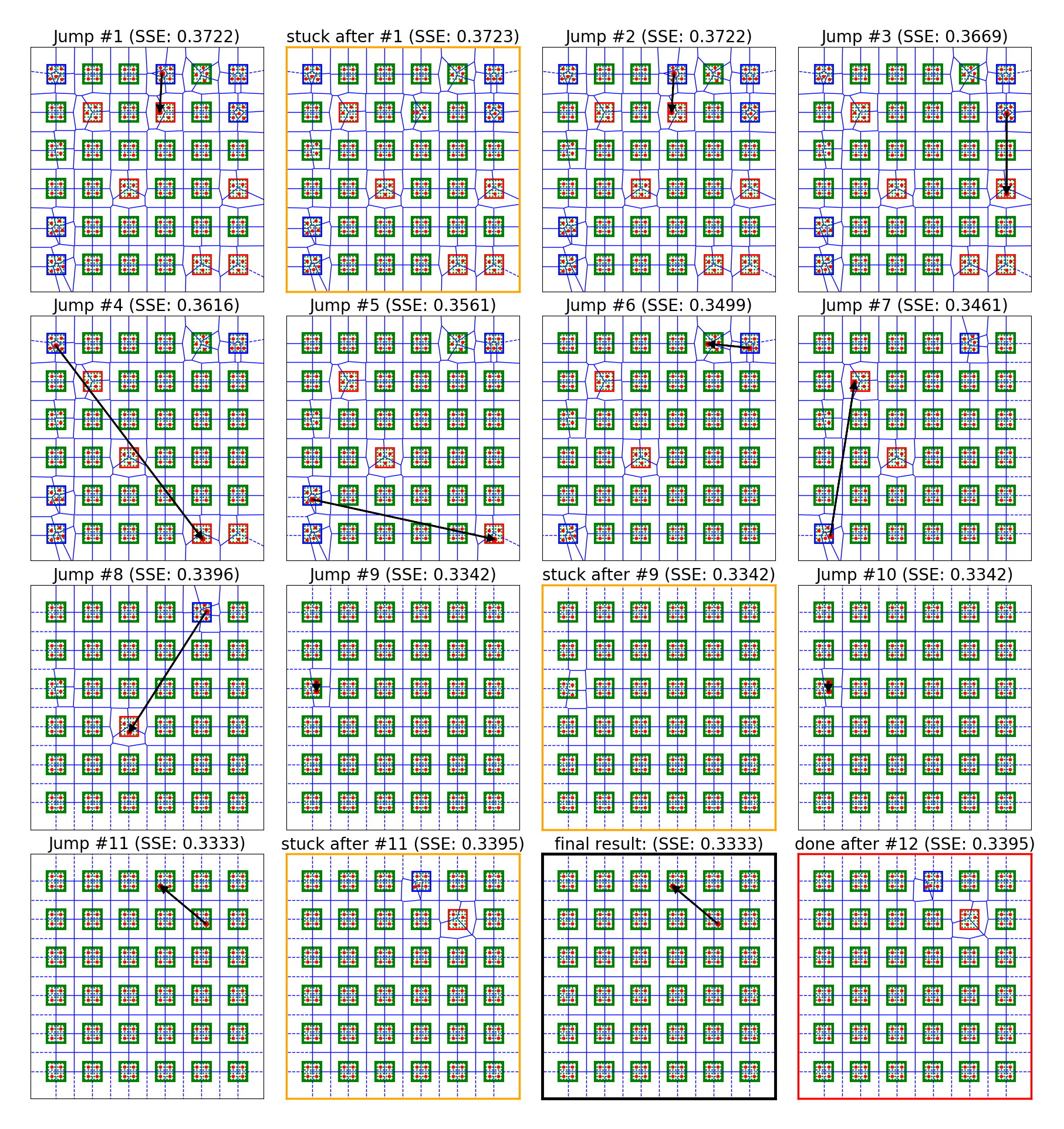}
	\caption[short caption]{\kms{} performs non-local jumps. We allow one retry per error level ($retry_{max} = 1$). Already after jump \#1 the error increases, so \kmu{} would have stopped right there (resulting SSE: 0.3722). We write "stuck after ..." above those cases and frame them in orange. The problematic constellation is actually very similar to the one shown in \cref{fig:arrowspoordetail} b). The performed retry luckily results in a lower error and the \kmu{} algorithm can continue for 6 successful jumps (reaching an SSE of 0.3342). Jump \#9 leads to an error increase again but also in this case the retry is successful and directly leads to the final solution with an SSE of 0.3333 which happens to be the optimum for this particular clustering problem. Further jumps and retries cannot improve this result, so \kms{} terminates shortly after. From the definition of the algorithm it follows that every \kms{} simulation must end with a sequence of $retry_{max}$ unsuccessful retries (one in this case). }
	\label{fig:arrowskms}
\end{figure}

In \cref{fig:arrowskms} a typical simulation sequence of \kms{} is shown. While \kmu{} would have stopped after 7 jumps, \kms{} continues, in this case to the optimum.

\section{Empirical results}\label{sec:empirical}

We performed systematic tests of \kmp, \kmu{} and \kms{} with 5 different data sets. For each data set a large range of values for $k$ was investigated. For each of these $k$-values 10 different simulation runs were performed. Each single simulation consisted of three phases:
\begin{enumerate}
	\item \kmp{} (i.e.\, the \kmp{} seeding followed by \km)
	\item \kmu, starting from the result of the \kmp{} run and continuing until the SSE did not fall anymore (the stopping criterion of \kmu)
	\item \kms{} starting from the result of \kmu{} and allowing 2 retries for each time \kmu{} came to a stop.
\end{enumerate}

\begin{figure}[p]
	\centering
	\includegraphics[width=0.50\linewidth]{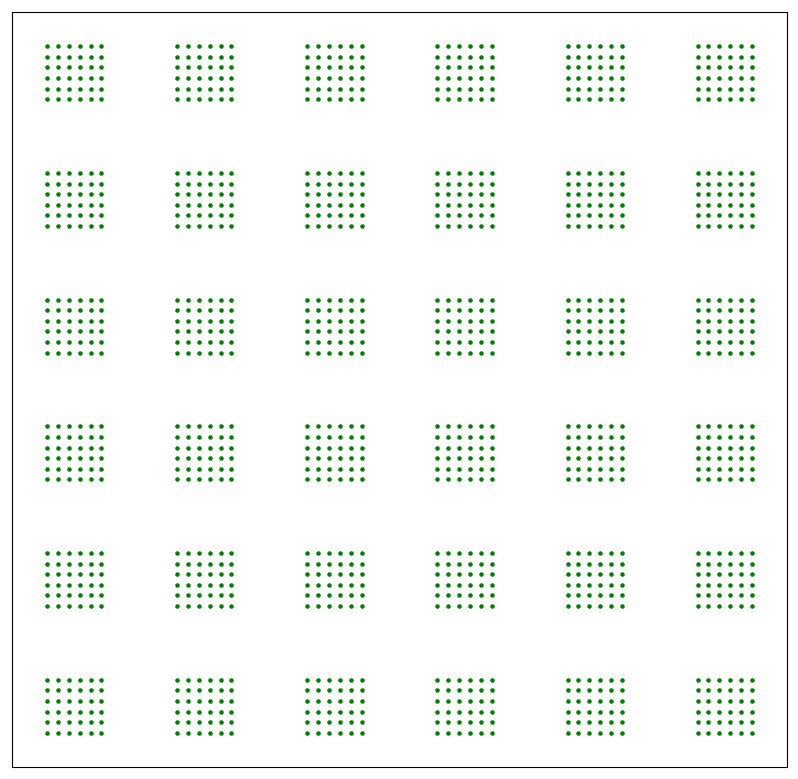}
	\caption[short caption]{Data set $A$ from \cref{fig:kma1_opt}, dimensionality $d=2$, number of data points $n=1296$ ($=36*36$), number of clusters $g=36$.}
	\label{fig:figpaperdata}
\end{figure}

\begin{figure}[p]
	\centering
	\includegraphics[width=0.95\linewidth]{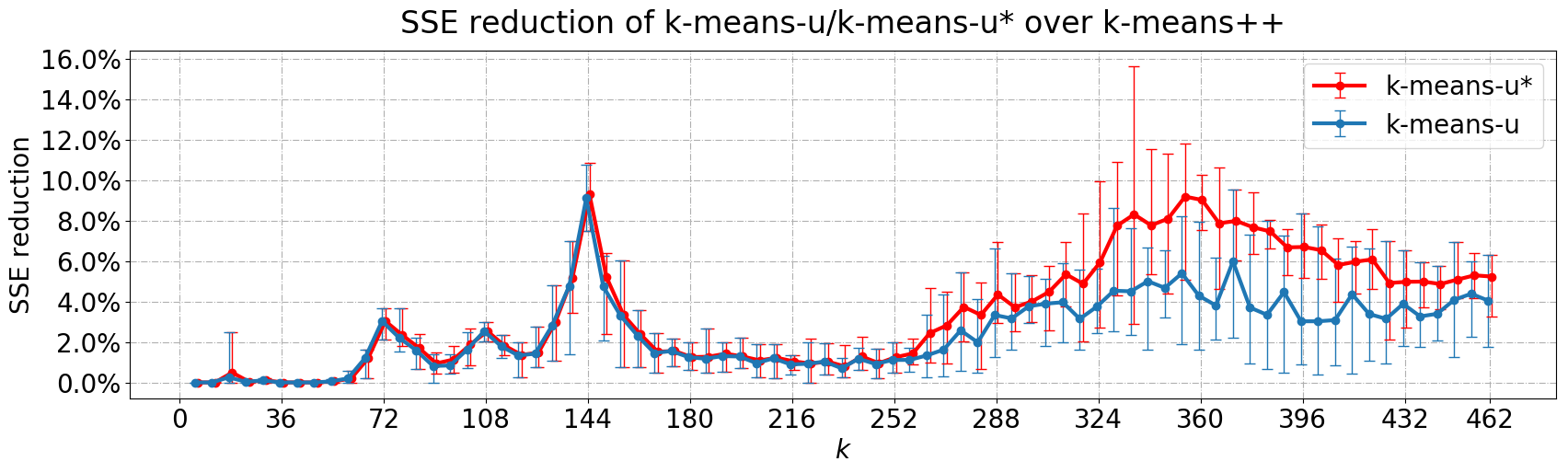}
	\caption[short caption]{Simulation results for data set $A$ (see \cref{fig:figpaperdata}).  In accordance with our analysis \kmp{} finds good results until $k=36$ but solution quality degrades (shown by the large improvements of \kmu{} and \kms) if $k$ is increased to $72$ or further multiples of 36. The problem $A$-4 illustrated in \cref{fig:kmppa4,fig:kmua4,fig:arrows,fig:arrowspoor,fig:arrowspoordetail,fig:arrowskms}  corresponds to $k=144$. For values of $k>250$ the improvements obtained by \kmu{} seem to grow independently of $k$ being an integral multiple of 36 and the effect of adding a greedy search (\kms) becomes very prominent, occasionally doubling the already significant improvements obtained by \kmu. 10 runs per $k$-value}
	\label{fig:figgaussian2Dstat}
\end{figure}

\begin{figure}[htp]
	\centering
	\includegraphics[width=0.50\linewidth]{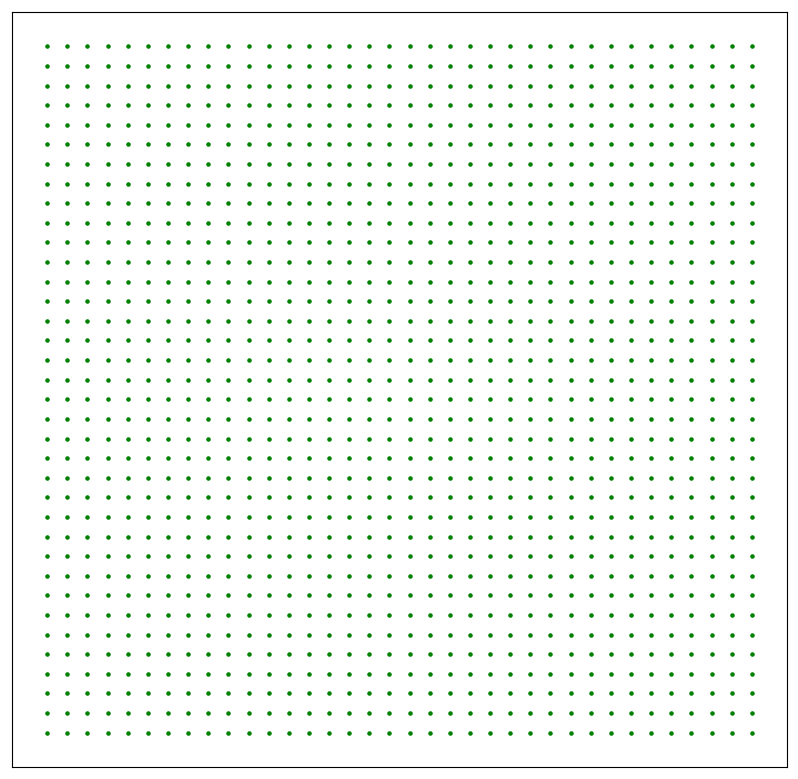}
	\caption[short caption]{Data set $B$. This dataset which has exactly the same number of points as dataset $A$  has been included as an example of a very simple unstructured data set with the expectation that there would be no large improvements of the \kmp{} results by our algorithms, a wrong assumption as the simulation results (see \cref{fig:figflat2Dstat}) show. dimensionality $d=2$, number of data points $n=1296$ ($=36*36$), number of clusters $g=1$ (or $g=1296$ depending on interpretation).}
	\label{fig:figflatdata}
\end{figure}

\begin{figure}[htp]
	\centering
	\includegraphics[width=0.95\linewidth]{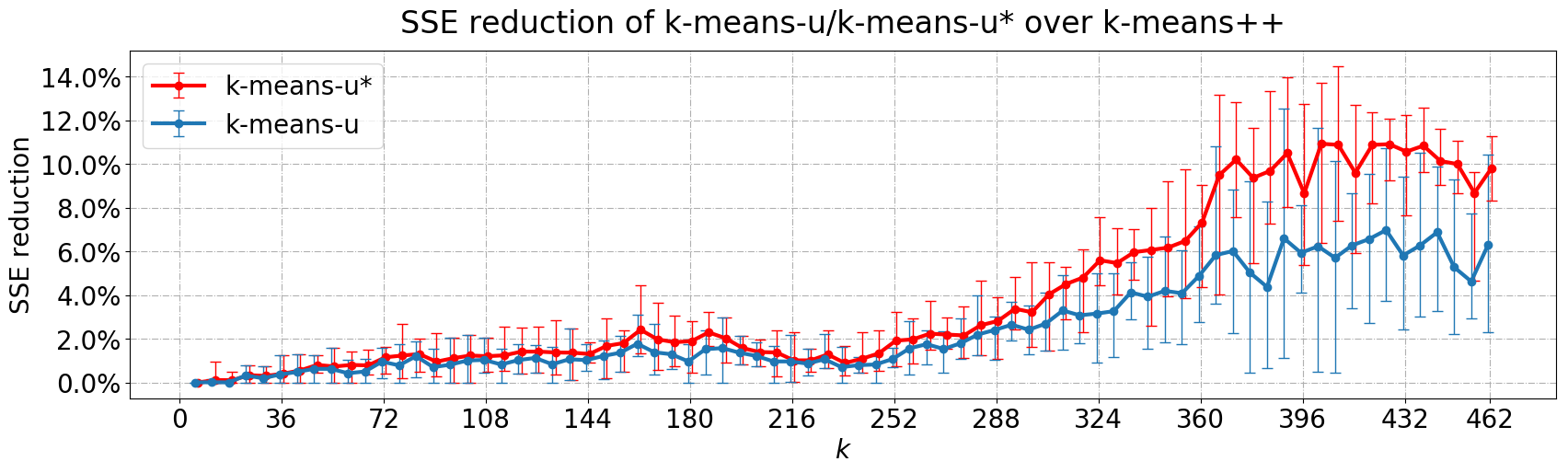}
	\caption[short caption]{Simulation results for data set $B$ (see \cref{fig:figflatdata}).   For smaller values of $k$ the improvements over \kmp{} are moderate (up to 2\%) and \kms{} does not deliver a large advantage over \kmu{}. Starting approximately with $k=288$ and increasingly with larger $k$-values however, \kmu{} is able to find solutions up to 6\% better (in the mean) than \kmp{} and \kms{} even finds solutions up to 11\% better (in the mean) than those of \kmp. 10 runs per $k$-value}
	\label{fig:figflat2Dstat}
\end{figure}

\begin{figure}
	\centering
	\includegraphics[width=0.60\linewidth]{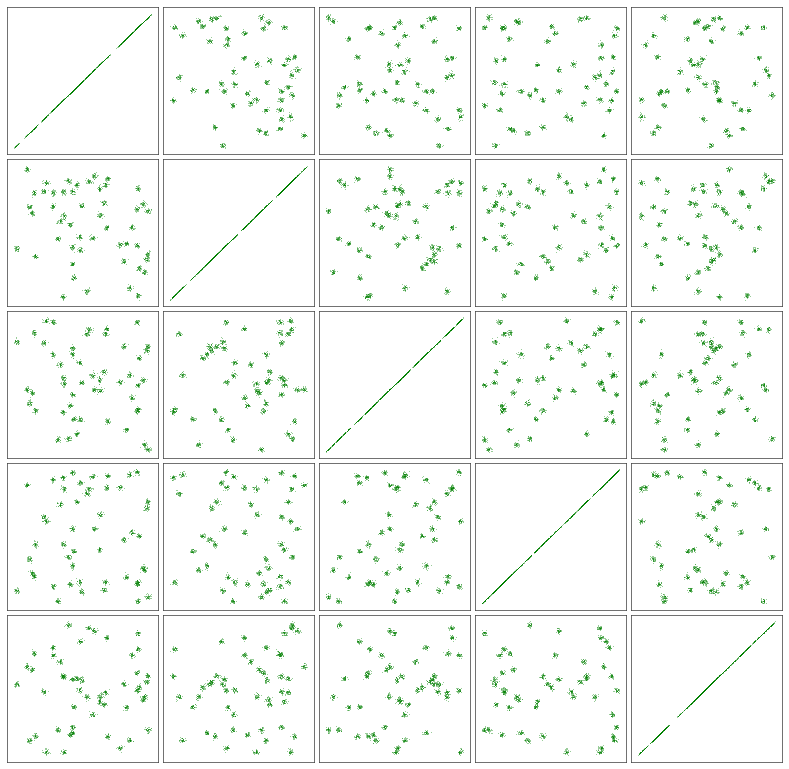}
	\caption[short caption]{5-dim.~mixture of 50 Gaussians (all pairs of dimensions displayed),  normally distributed  with $\sigma=0.00001$ and unit covariance matrix,  $d=5$, $n=2000$, $g=50$}
	\label{fig:figgaussian5D}
\end{figure}

\begin{figure}
	\centering
	\includegraphics[width=0.95\linewidth]{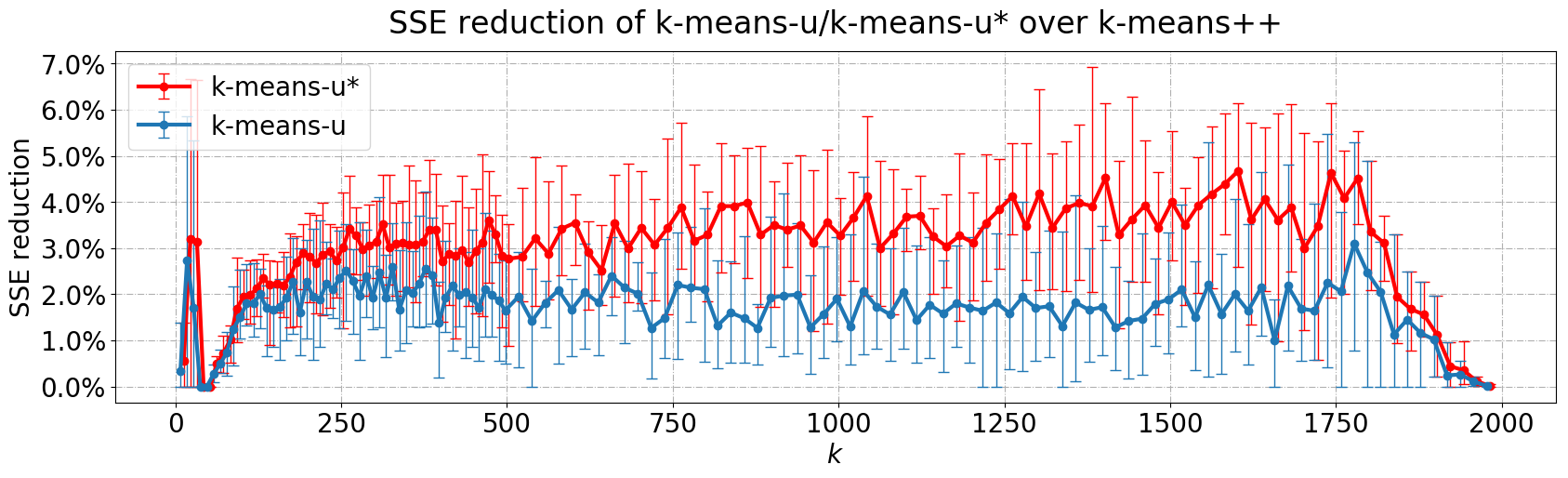}
	\caption[short caption]{Simulation results for 5-dim. mixture of 50 Gaussians (see \cref{fig:figgaussian5D}). One can note that both \kmu{} and \kms{} are unable to find improvements for $k=50$ (50 is also the number of clusters in the data set) but for some smaller values of $k$ for all larger values considerable improvements are found. 10 runs per $k$-value}
	\label{fig:figgaussian5Dstat}
\end{figure}

\begin{figure}
	\centering
	\includegraphics[width=0.89\linewidth]{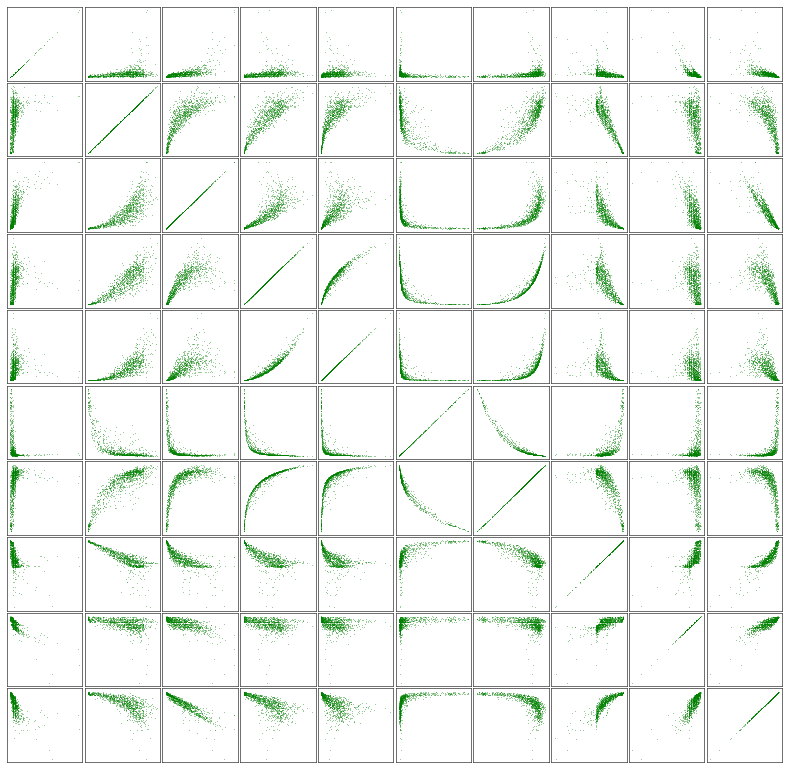}
	\caption[short caption]{Cloud data from UCI (https://archive.ics.uci.edu/ml/data sets/Cloud), (all pairs of dimensions displayed), data preprocessed with {\ttfamily sklearn.preprocessing.StandardScaler} to have unit variance in each direction, $d=10$, $n=1024$, $g=$(unknown)
	}
	\label{fig:clouddata}
\end{figure}

\begin{figure}
	\centering
	\includegraphics[width=0.95\linewidth]{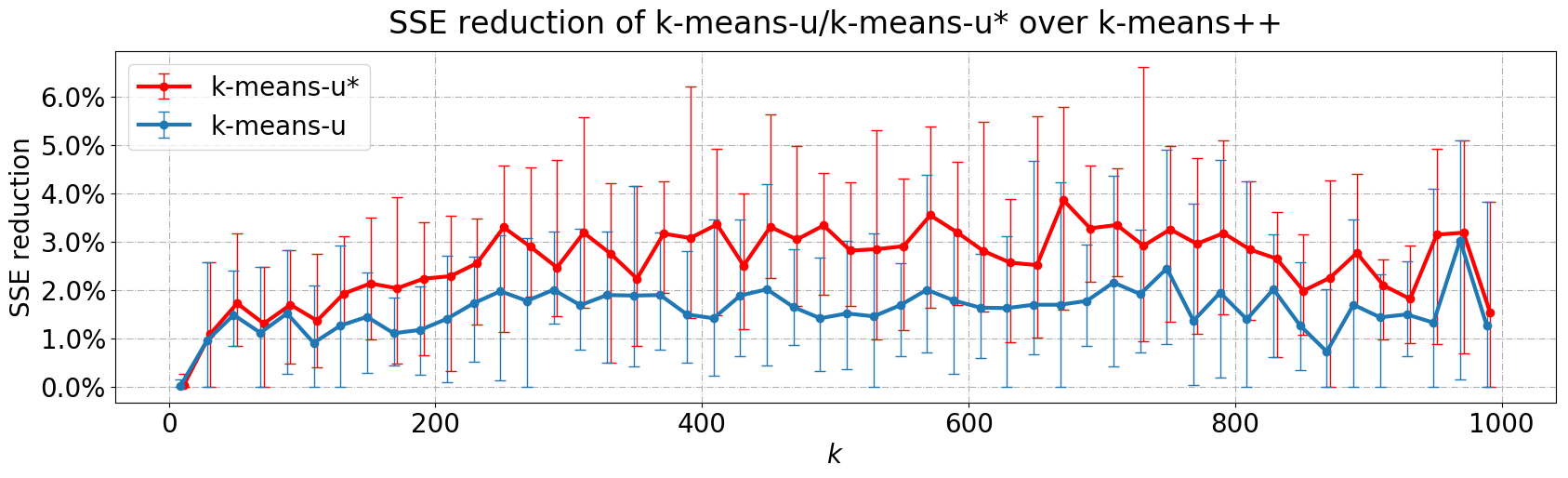}
	\caption[short caption]{Simulation results for Cloud data from UCI (see \cref{fig:clouddata}), 10 runs per $k$-value}
	\label{fig:cloudstat}
\end{figure}

\begin{figure}
	\centering
	\includegraphics[width=0.89\linewidth]{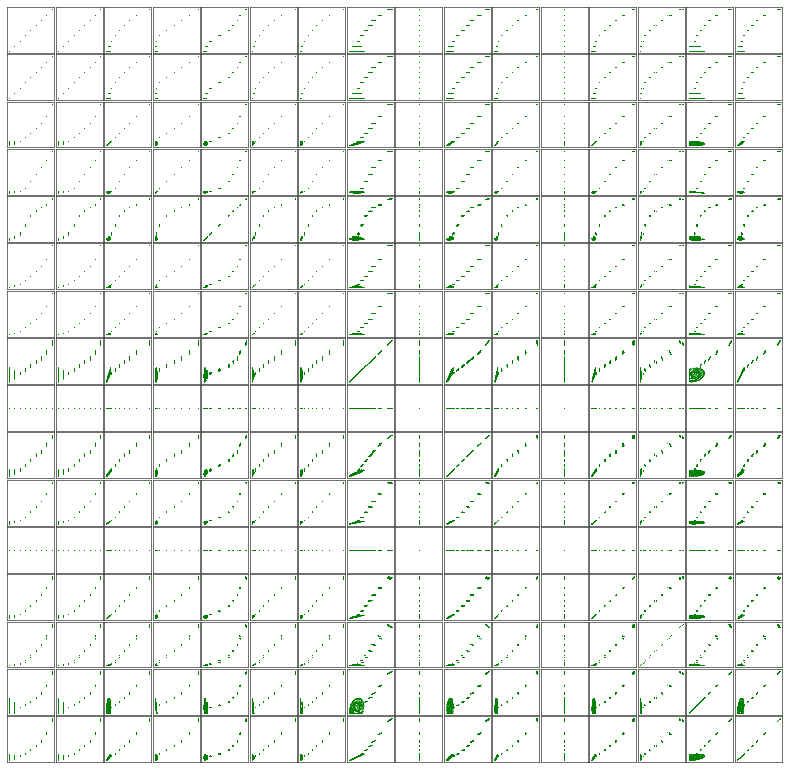}
	\caption[short caption]{Propulsion data from UCI (https://archive.ics.uci.edu/ml/data sets/Condition+Based+Maintenance+of+Naval+Propulsion+Plants), (all pairs of dimensions displayed), data preprocessed with {\ttfamily sklearn.preprocessing.StandardScaler} to have unit variance in each direction,  $d=16$, $n=11934$, $g=$(unknown)
	}
	\label{fig:propulsiondata}
\end{figure}

\begin{figure}
	\centering
	\includegraphics[width=0.95\linewidth]{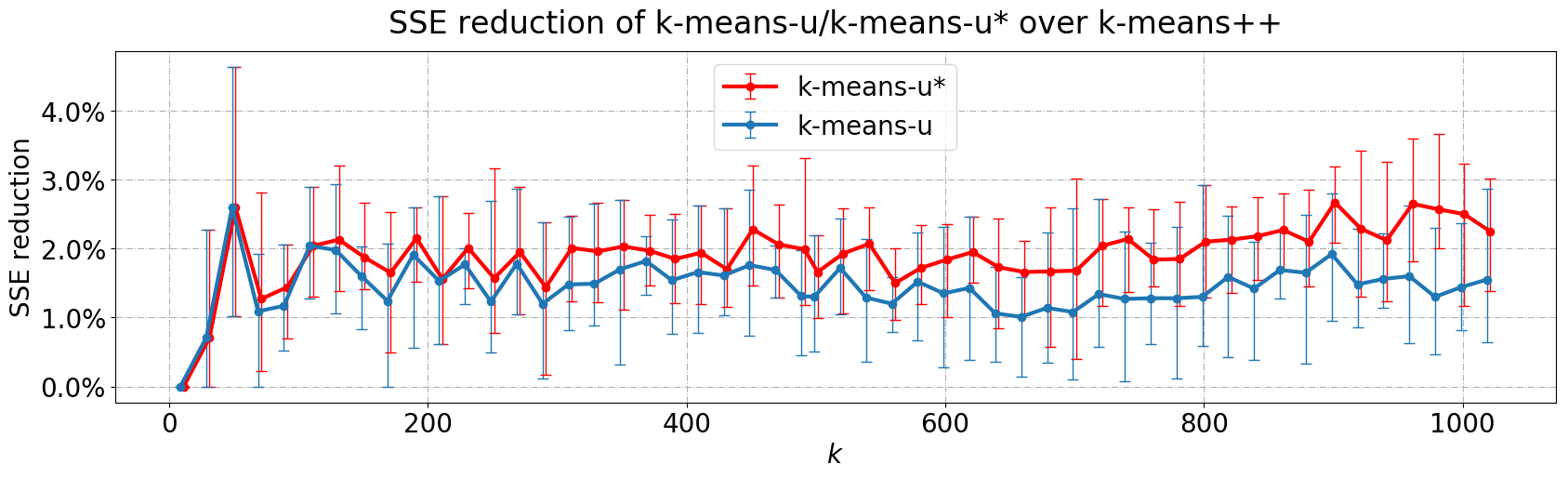}
	\caption[short caption]{Simulation results for propulsion data from UCI (see \cref{fig:propulsiondata}),  10 runs per $k$-value}
	\label{fig:propulsionstat}
\end{figure}

For each data set we show an illustration of the data set itself and a performance chart (\crefrange{fig:figpaperdata}{fig:propulsionstat}). If the dimension $d$ of the data set is larger than two, we display all $d^2$ pairs of dimensions, each in a separate sub plot. The scaling of each sub plot is chosen such that the whole available area is used to display data points. Therefore different subplots may have different scalings, but the general nature of the data should be more visible this way.

The performance chart takes the performance of \kmp{} as the baseline and indicates for both \kmu{} and \kms{} by how many percent they did reduce the SSE obtained by \kmp. No improvement would correspond to a data point on the $k$-axis and any actual improvement to data points above the $k$-axis. Per construction the new algorithms can not deliver anything worse than \kmp{} so there are no values below the $k$-axis. For both \kmu{} and \kms{} the mean improvement (main chart) as well as the minimum and maximum improvements (error bars)  are shown. While the mean indicates what to expect from one algorithm run, the maximum is an indication of what one could achieve by picking the best result of several runs.

The figure captions of the performance charts contain specific remarks regarding the simulation results. In general the new algorithms were able to improve a clear majority  of the \kmp{} results and often by a large margin. \kms{} in particular was not only able to raise the mean improvement compared to \kmu{} but also to obtain in many cases much higher maximum values (often more than 2 times as high as the maximum values of \kmu). 

With the exception of data set $A$ (\cref*{fig:figpaperdata}) which - as we know from earlier sections - is challenging for \kmp and data set $B$ (\cref{fig:figflatdata})which was included as an example of an unstructured data set the other data sets have not been constructed or chosen with any result in mind but rather to provide a certain variety. Two data sets (cloud and propulsion) were taken from the UCI Machine Learning Repository as an established source of well-kept data sets. The simulations were performed in python using the optimized implementation of \kmp{}  contained in the \texttt{scikit-learn} package and a (non-optimized) numpy-based implementation of \kmu{} and \kms. Since on this base the comparison of running times was difficult, we compared the number of Lloyd iterations. 

\begin{figure}
	\centering
	\includegraphics[width=0.95\linewidth]{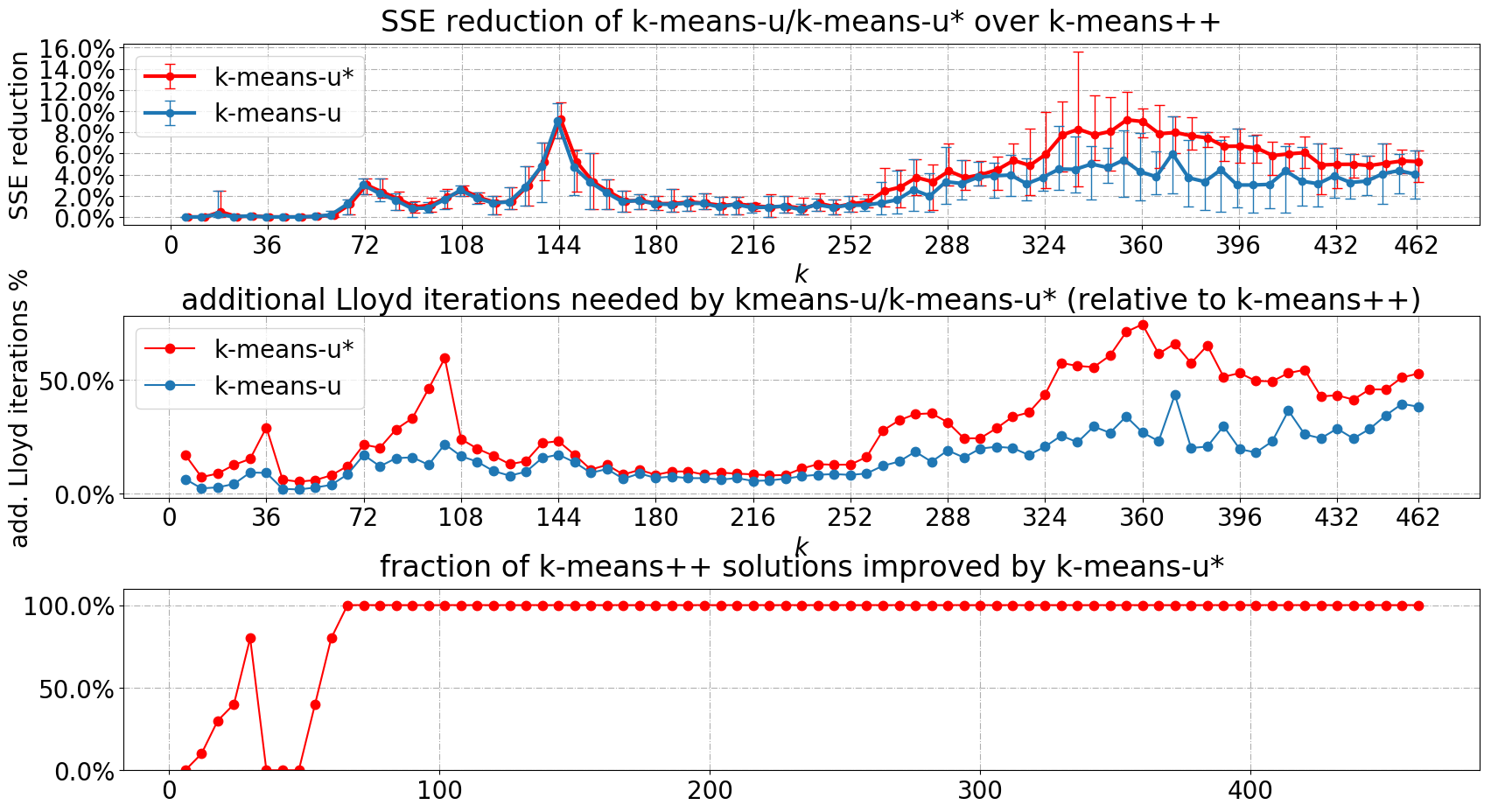}
	\caption[short caption]{Data set $A$: SSE reduction achieved (top), additional Lloyd iterations needed (center), and fraction of \kmp{} solutions improved by \kms{} (bottom). For large values of $k$ there seems to be a correlation between achieved SSE reduction and additional Lloyd iterations needed. }
	\label{fig:gaussian2DstatZZZ}
\end{figure}

\begin{figure}
	\centering
	\includegraphics[width=0.95\linewidth]{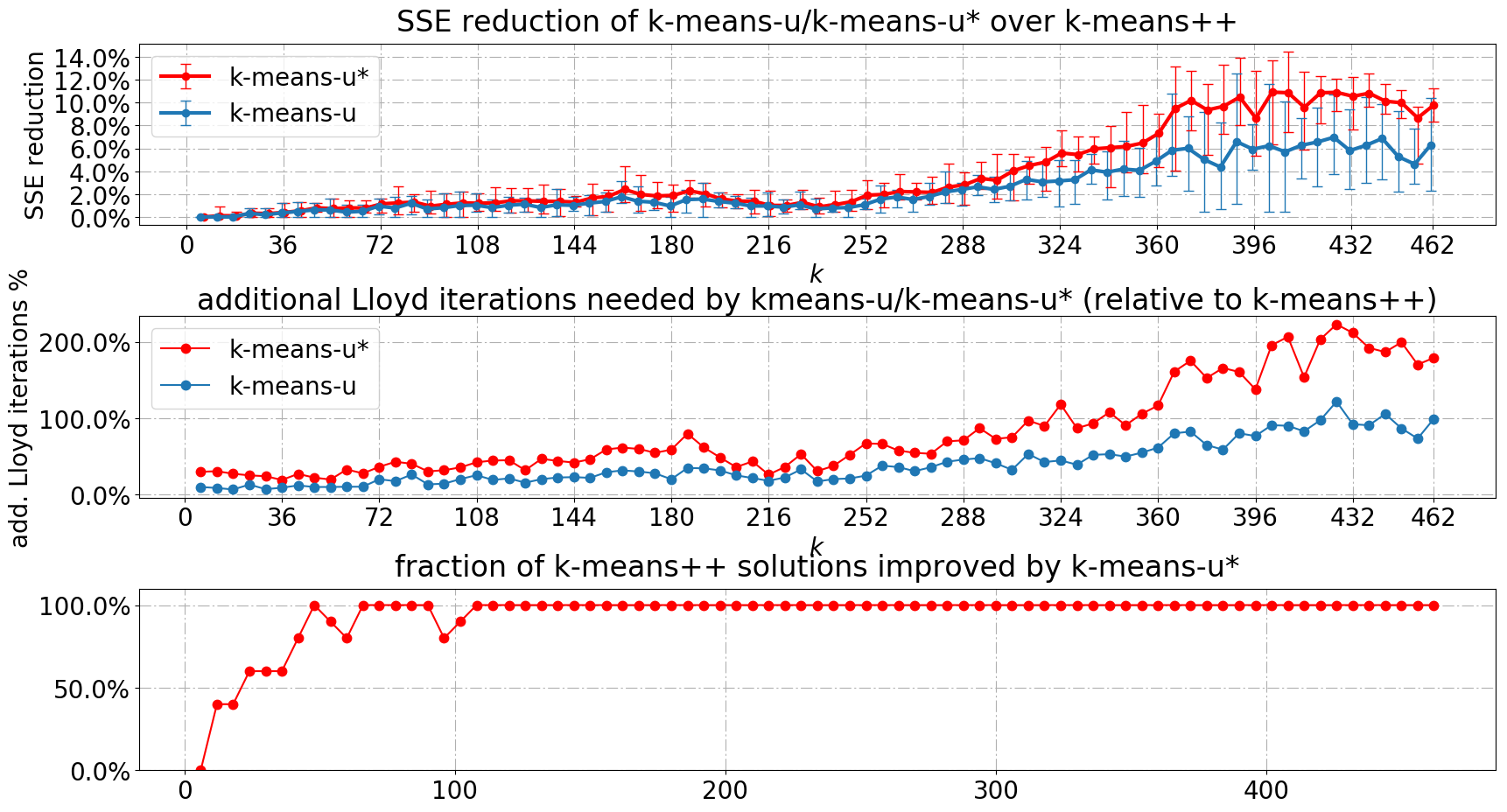}
	\caption[short caption]{Data set $B$ from \cref{fig:figflatdata}: SSE reduction achieved (top), additional Lloyd iterations needed (center), and fraction of \kmp{} solutions improved by \kms{} (bottom). One can not that the curves are similar to those for dataset $A$ (\cref{fig:gaussian2DstatZZZ}) but lack the peaks where $k$ is a low multiple of 36. The correlation between achieved SSE reduction and additional Lloyd iterations needed seems to be present for the whole range of $k$.}
	\label{fig:flat2DstatZZZ}
\end{figure}

\begin{figure}
	\centering
	\includegraphics[width=0.95\linewidth]{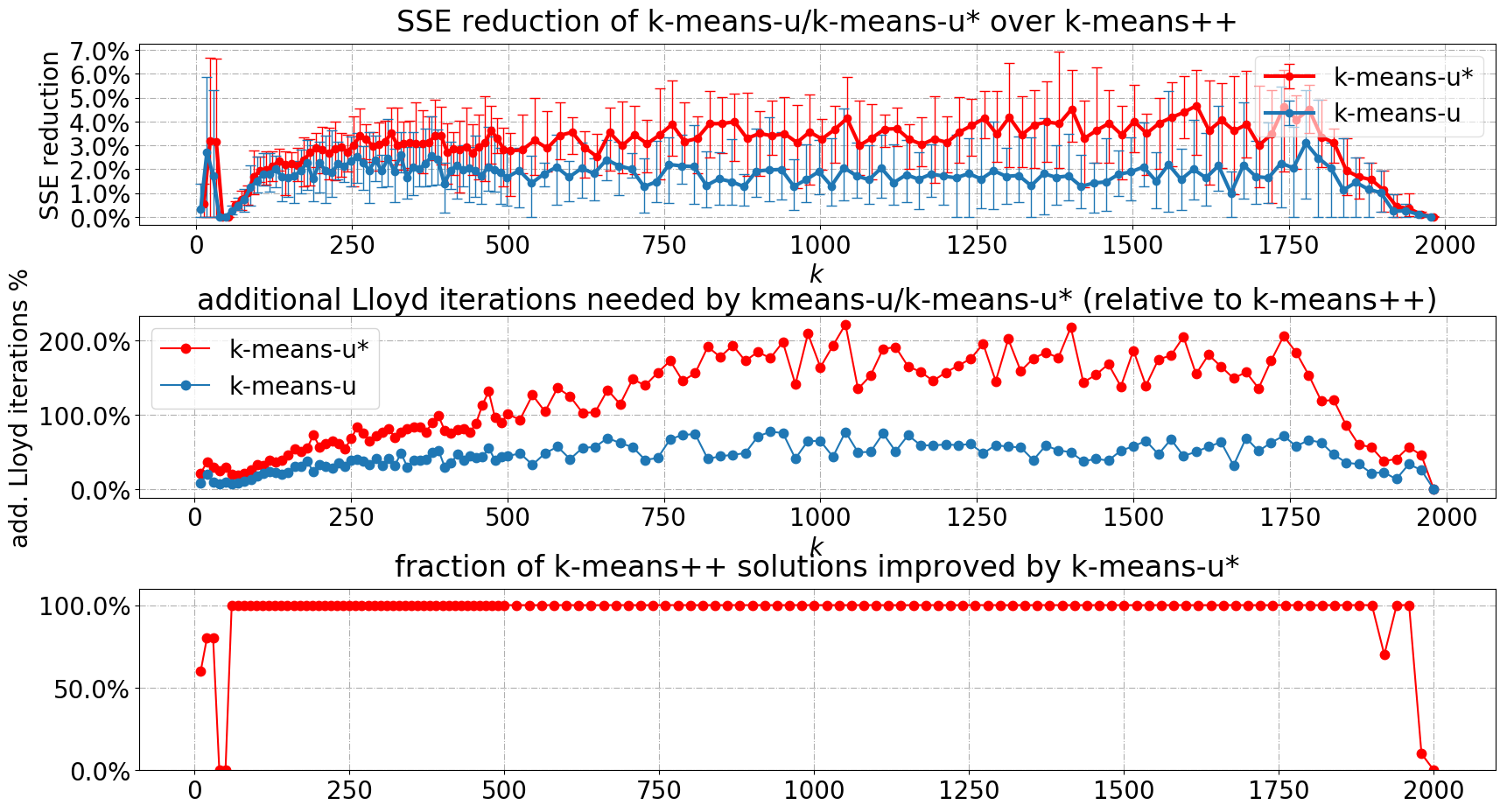}
	\caption[short caption]{5-dim Mixture of Gaussians: SSE reduction achieved (top), additional Lloyd iterations needed (center), and fraction of \kmp{} solutions improved by \kms{} (bottom). For this problem the computation overhead compared to \kmp{} is particularly large. \kms{} achieves an improvement around 3\% over \kmp{} in most cases.}
	\label{fig:gaussian5DstatZZZ}
\end{figure}

\begin{figure}
	\centering
	\includegraphics[width=0.95\linewidth]{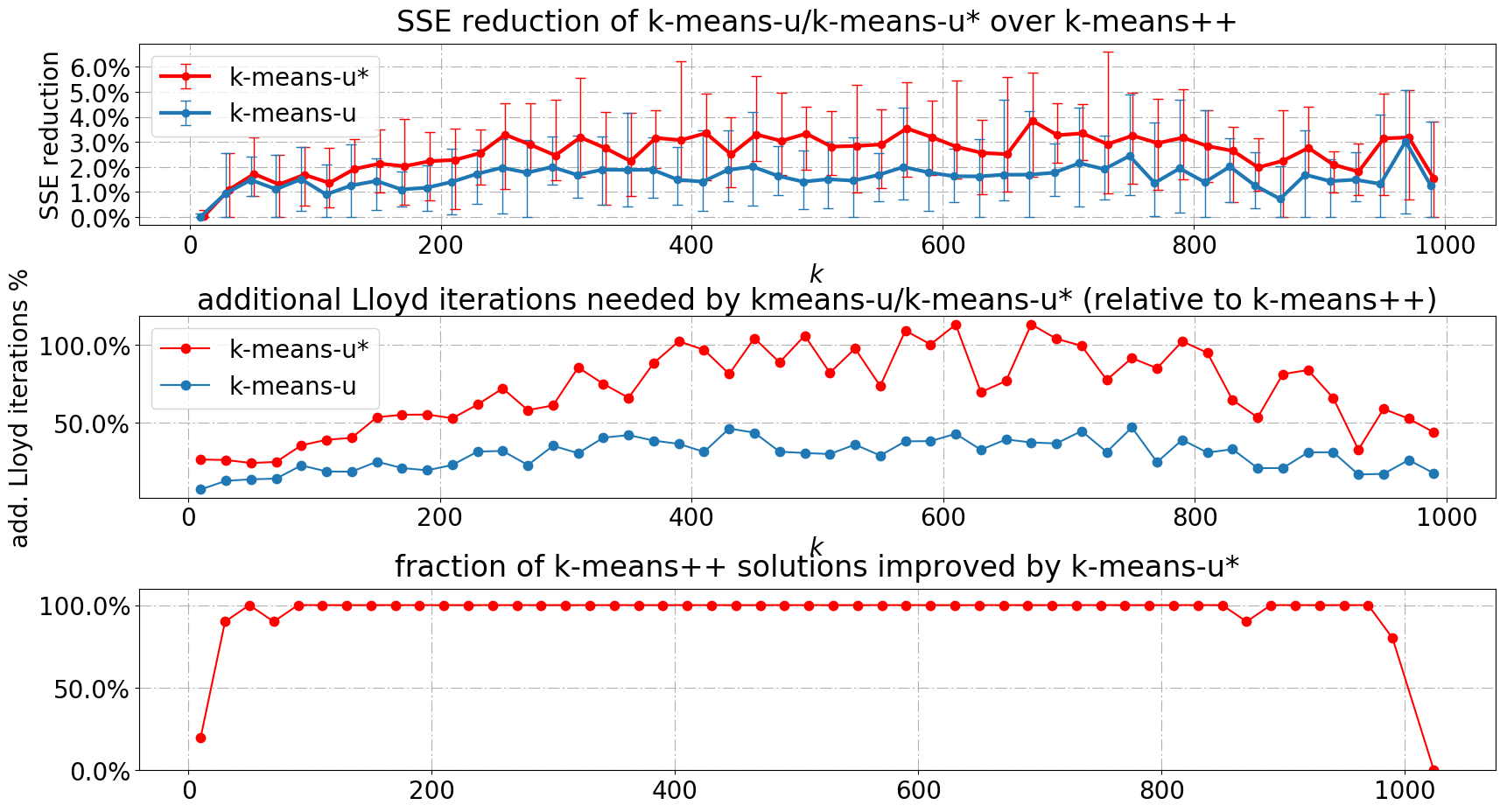}
	\caption[short caption]{Cloud data: SSE reduction achieved (top), additional Lloyd iterations needed (center), and fraction of \kmp{} solutions improved by \kms{} (bottom).}
	\label{fig:cloudstatZZZ}
\end{figure}

\begin{figure}
	\centering
	\includegraphics[width=0.95\linewidth]{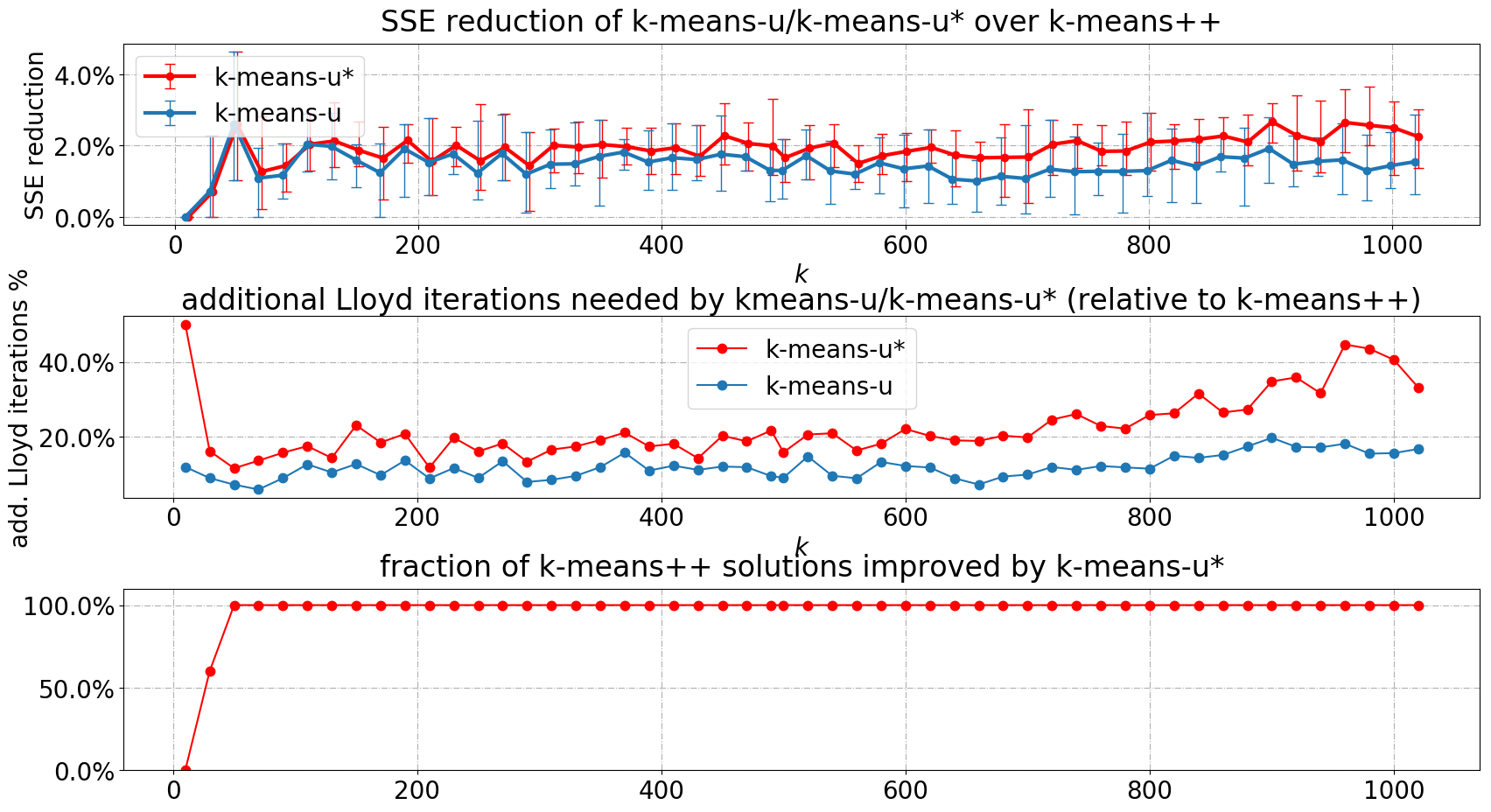}
	\caption[short caption]{Propulsion data: SSE reduction achieved (top), additional Lloyd iterations needed (center), and fraction of \kmp{} solutions improved by \kms{} (bottom). One can note that both the SSE improvements over \kmp{} (mostly around 2\%) as well as the additionally needed Lloyd iterations compared to \km{} (less than 50\%, mostly even below 25\%) are quite low. One could speculate that for this particular data set \kmp{} already finds quite good solutions which are difficult to improve. Nevertheless - as the bottom chart shows -  starting with $k=50$ every single solution found by \kmp{} is improved by \kms.}
	\label{fig:propulsionstatsZZZ}
\end{figure}

In \crefrange{fig:gaussian2DstatZZZ}{fig:propulsionstatsZZZ} we display for all performed simulations the relative overhead of \kmu{} and \kms{} in terms of Lloyd iterations  as well as the fraction of \kmp{} solutions which \kms{} was able to improve. The error reduction shown earlier is repeated for reference as well. 

The overhead of \kms{} over \kmp{} measured as described ranged between 10\% and 230\%. In computing this we considered that \kmp{} is executed 10 times in \texttt{scikit-learn} before the best result is returned. This means that for all our experiments the effort for the more complex one of our algorithms (\kms) was within a constant factor (3.3) of \kmp. Given that \kms{}  provides significant solution improvements for an NP-complete problem this can be seen as a very moderate effort.

With few exceptions (mainly for small values of $k$ or values of $k$ near the number $n$ of data points) the percentage of \kmp{} solutions improved by \kms{} is at 100\%. Thus according to our experiments it is highly likely that an arbitrary solution found by \kmp{} can be further improved by \kms{}.




\clearpage
\section{Theoretical Bound for Solution Quality}\label{sec:guarantees}

No specific formal analysis of the new algorithm has been performed so far. However, \kms{} starts off from the result of \kmp{} and does only deliver a different result, if it is better than that of \kmp. Therefore (and trivially so) the same bound which has been established for \kmp{} holds for the new algorithm as well:

\begin{equation}
E[\phi] \le 8(\mbox{ln}\,k+2)\phi_{\mbox{OPT}}
\end{equation}
So it is guaranteed that the summed squared error of any solution computed by \kms{} differs from the optimum only by a factor which is logarithmically dependent on $k$. The above bound holds for \kmu{} as well, since it also returns the initial \kmp{} solution if it cannot produce a better one.

\section{Summary}\label{sec:summary}

In this paper we propose  the \kms{} algorithm. It is based on the initially proposed \kmu{} algorithm which performs non-local jumps based on a simple utility criterion to improve the clustering results obtained with the current de-facto standard method \kmp. 
In some cases however we observed \kmu{} stopping too early due to an error increase caused by a poor local minimum in which the \km{} phase of \kmu{} ended.  This behavior was largely overcome by allowing a small and finite number of retries  (e.g. 2) for the most recent jump. Due to the randomized local positioning during a jump, the resulting configuration after a retry is often different and possibly leads to a lower error which allows the algorithm to continue with the next jump. Further retries are then possible, but only on lower error levels which leads to a strictly monotone decreasing sequence of error values until the algorithm terminates. The resulting extended version of \kmu{} is called the \kms{} algorithm. By construction the logarithmic quality bound established for \kmp{} holds for \kms{} as well. 

Simulations with a variety of data sets (partially from the UCI Machine Learning Repository) demonstrate that \kmp{} is dominated w.r.t.~solution quality (SSE) by \kmu{} which from a certain value of $k$ very often generates significantly better solutions and that \kmu{} itself is dominated by \kms{},  again significantly in a large number of cases. The observed improvements over \kmp{} depended on the structure of the data distribution and the number $k$ of centers and ranged from zero to about 8\% mean reduction of the summed squared error. Our method incurs only a moderate computational overhead compared to \kmp{}  (below 230\% in all of our experiments, less than 50\% for the propulsion data, the largest of the data sets we used). Since the problem of minimizing SSE is NP-complete for data dimensions of two and larger, these additional costs for achieving often several percent error reduction appear to be quite reasonable.
In conclusion we consider \kms{} a potential replacement of \kmp{} in all cases where the quality of the clustering is of high importance.

\appendix
\section{Example Implementation}
The complete python code used for the experiments in this paper is available from\\ {\ttfamily https://github.com/gittar/k-means-u-star}.


\vskip 0.2in
\bibliography{kmeansu}

\end{document}